\documentclass[10pt,twocolumn,letterpaper]{article}

\usepackage{iccv}
\usepackage{times}
\usepackage{epsfig}
\usepackage{graphicx}
\usepackage{amsmath}
\usepackage{amssymb}

\usepackage{bm}
\usepackage{booktabs}

\usepackage[font=small]{caption}
\usepackage[caption=false]{subfig}
\usepackage{multirow}
\usepackage[linesnumbered,ruled,vlined]{algorithm2e}
\SetKwComment{Comment}{/* }{ */}
\usepackage{adjustbox}

\usepackage[accsupp]{axessibility}

\usepackage[breaklinks=true,colorlinks,bookmarks=false]{hyperref}

\iccvfinalcopy 

\def\iccvPaperID{11128} 

\ificcvfinal\fi

\begin{document}

\title{CGBA: Curvature-aware Geometric Black-box Attack}

\author{Md Farhamdur Reza, Ali Rahmati, Tianfu Wu, and Huaiyu Dai\\ \\
Department of ECE, North Carolina State University \\ 
{\tt\small mreza2@ncsu.edu, arahmat@alumni.ncsu.edu, tianfu\_wu@ncsu.edu , hdai@ncsu.edu}}

\maketitle
\ificcvfinal\thispagestyle{empty}\fi

\begin{abstract}
Decision-based black-box attacks often necessitate a large number of queries to craft an adversarial example. 
Moreover, decision-based attacks based on querying boundary points in the estimated normal vector direction often suffer from inefficiency and convergence issues. 
In this paper, we propose a novel query-efficient \b curvature-aware \b geometric decision-based \b black-box \b attack (CGBA) that conducts boundary search along a semicircular path on a restricted 2D plane to ensure finding a  boundary point successfully irrespective of the boundary curvature.  
While the proposed CGBA attack can work effectively for an arbitrary decision boundary, it is particularly efficient in exploiting the low curvature to craft high-quality adversarial examples, which is widely seen and experimentally verified in commonly used classifiers under non-targeted attacks. In contrast, the decision boundaries often exhibit higher curvature under targeted attacks.
Thus, we develop a new query-efficient variant, CGBA-H, that is adapted for the targeted attack. 
In addition, we further design an algorithm to obtain a better initial boundary point at the expense of some extra queries, which considerably enhances the performance of the targeted attack. Extensive experiments are conducted to evaluate the performance of our proposed methods against some well-known classifiers on the ImageNet and CIFAR10 datasets, demonstrating the superiority of CGBA and CGBA-H over state-of-the-art non-targeted and targeted attacks, respectively. 
The source code is available at \url{https://github.com/Farhamdur/CGBA}.
\end{abstract}

\section{Introduction}
\label{sec:intro}
Adversarial attacks are broadly classified into two types: white-box and black-box attacks. In the white-box attack setting~\cite{goodfellow2014explaining, moosavi2016deepfool, carlini2017towards}, the adversary possesses the full knowledge of the target classifier and its weights. 
However, it is often impractical to avail information about the target classifier in real-world scenarios. Therefore, the black-box setting—transfer-based, score-based, and decision-based—is the practical setting for adversarial attacks with limited knowledge of the classifier.  The transfer-based adversarial attacks~\cite{papernot2016transferability, wu2020skipTransferability} use a surrogate model to generate adversarial examples though it does not guarantee a high attack success rate. Score-based attacks~\cite{Scorechen2017zoo, Scoreilyas2018prior} query the target classifier for the prediction probabilities of all the classes in order to estimate the gradient in each step and lessen perturbation. However, this attacking strategy may not be feasible because, in many real-world applications, classifiers only return the top-1 classification label in response to a query. 
Thus, the decision-based adversarial attack is the most practical adversarial attack as it allows the adversary to craft an adversarial example by only querying the top-1 classification label from the target classifier.

Most state-of-the-art decision-based attacks, such as HSJA~\cite{chen2019hopskipjumpattack}, qFool~\cite{liu2019qFool}, GeoDA~\cite{geoda}, QEBA~\cite{li2020qeba} and TA~\cite{ma2021tangent}, are based on finding the normal vector at a point on the decision boundary and iteratively search for new boundary points with reduced perturbation. Among these attacks, HSJA~\cite{chen2019hopskipjumpattack}, qFool (targeted)~\cite{liu2019qFool}, QEBA~\cite{li2020qeba}, and TA~\cite{ma2021tangent} employ the estimated normal vector direction to obtain a point inside the adversarial region, and then apply binary search between the obtained adversarial point and the source to get a new boundary point. 
The aforementioned approaches, however, do not explicitly take into account the geometry of the boundary when coining adversarial examples. qFool (non-targeted)~\cite{liu2019qFool} and GeoDA~\cite{geoda}, on the other hand, approximate the boundary as a hyperplane and find a new boundary point by conducting a \b binary \b search along the direction of the estimated \b normal \b vector (BSNV). 
As further discussed below, while BSNV is effective for low-curvature decision boundaries where linear approximation is sufficiently accurate, its effectiveness deteriorates with the increase of curvature and nonlinearity at the decision boundaries. For a boundary with high curvature, BSNV may not even hit the boundary due to the narrow adversarial region.
SurFree~\cite{maho2021surfree} also considers a hyperplane boundary and conducts boundary search along a semicircular path, but it does not utilize the information of the normal vector and instead estimates the attack direction through random trials.

A careful examination of the above normal vector based attacks reveals the following limitations. The estimation of the normal vector may be inaccurate due to the limited query budget and the non-linearity of the boundary. Thus, the expected reduction in perturbation may not occur when searching along the direction of the estimated normal vector. 
Moreover, if the adversarial region is narrow enough, the search process does not converge towards perturbation reduction due to the inability to find the adversarial region in the search direction.
Fundamentally, these limitations are related to the one-dimensional (1-D) search nature dictated by the estimated normal vector. 
The state-of-the-art (SOTA) non-targeted attack SurFree, on the other hand, queries an adversarial point along a semicircular path but does not use the critical normal vector information to estimate the attack direction.
Motivated by the above observation, we propose a new curvature-aware geometric black-box attack (CGBA) in this work to further improve the attack efficiency. Particularly, rather than conducting a boundary point search towards the estimated normal direction or along a semicircular path in some random direction, CGBA conducts the boundary point search along a semicircular path (BSSP) in a restricted 2-D plane spanned by two vectors: the direction towards a boundary point from the source (i.e., $\hat{\bm v}_t$ in Figure~\ref{al1_mainFig}) and the estimated normal direction on that boundary point (i.e., $\hat{\bm \eta}_t$ in Figure~\ref{al1_mainFig}). 
As further illustrated in Section~\ref{sect:proposed_methods} and Appendix~\ref{bsssp_bsnv}, the proposed CGBA overcomes the limitations of 1D boundary search and is a query-efficient approach for low curvature boundaries. However, it gradually loses query efficiency as the curvature of the boundary increases. Thus, we modify CGBA to CGBA-H which follows the same restricted 2D semicircular path but more swiftly adapts to the high curvature of the decision boundary. 
Our main contributions are summarized as follows:

\begin{itemize}
    
    \vspace{-1mm}
    \item We propose CGBA, a novel iterative decision-based black-box attack that conducts boundary search along a semicircular path on a restricted 2-D plane and effectively overcomes the limitations of existing 1-D search based on estimated normal vectors at the decision boundary.   
    \vspace{-1mm}
    \item The proposed CGBA attack can effectively exploit the decision boundary's low curvature for non-targeted attacks. When the decision boundary assumes a high curvature, we develop a new variant, CGBA-H, which achieves better performance for targeted attacks.  
    \vspace{-1mm}
    \item Moreover, we introduce an algorithm to choose a better initial boundary point and demonstrate that this initialization method leads to significant performance improvement for the targeted attack.  
    \vspace{-1mm}
    \item Experimental results on ImageNet and CIFAR10 reveal the efficacy of CGBA and CGBA-H for non-targeted and targeted attacks, respectively.
\end{itemize} 


\section{Related work}
Decision-based black-box attack is the most challenging setting to obtain adversarial examples as the only information available to perform this type of attack is the target classifier's top-1 classification label. Some decision-based black-box attacks use a random search, while others are based on finding the gradient on the decision boundary. Boundary Attack~\cite{brendel2018decision} algorithm performs a random walk along the decision boundary to reduce the perturbation with query though it still incurs a large number of queries. To speed up the performance of~\cite{brendel2018decision}, Biased Boundary Attack~\cite{brunner2019guessing} proposes three priors to reduce the search space, and it is shown that the Perlin bias introduces the most favorable effect.
OPT~\cite{cheng2018query} and Rays~\cite{chen2020rays} are decision-based attacks that randomly search for optimal directions to reduce the perturbation. However, Rays is only applicable for non-targeted attacks~\cite{ma2021tangent}, and its performance is shown for $\ell_{\infty}$-norm. Sign-OPT~\cite{cheng2020sign} improves the query efficiency of OPT~\cite{cheng2018query} by computing the sign of the directional derivatives to estimate the gradient.
In~\cite{dong2019evolutionary}, an evolutionary attack method is proposed in which random samples are drawn from a normal distribution with customized co-variance in reduced search space. AHA~\cite{li2021aha}, on the other hand,  utilizes the mean of the historical queries to generate random samples from a normal distribution. 
Triangle Attack (TriA)~\cite{wang2022Triangle} utilizes the geometry of a triangle to perform the decision-based attack.
SurFree~\cite{maho2021surfree}, a surrogate-free algorithm,  claims that bypassing the query cost of normal vector estimation would improve query efficiency. However, we refute this claim by conducting the boundary search in a restricted 2D plane guided by the normal vector and achieving better performance. Moreover, SurFree only supports non-targeted attacks \cite{ma2021tangent} as opposed to our methods addressing both non-targeted and targeted attacks.

Several existing attacks rely on estimating the normal vector on the boundary point. HSJA~\cite{chen2019hopskipjumpattack} proposes query-efficient methods by estimating the normal vector on the decision boundary to obtain a boundary point with reduced perturbation. qFool~\cite{liu2019qFool} and GeoDA~\cite{geoda} are based on the observation that the curvature of the decision boundary is small around adversarial examples. To improve the performance using the normal vector estimation, GeoDA~\cite{geoda}, which is applicable for non-targeted attacks, proposes a method to distribute the query optimally to iterations given a query budget. QEBA~\cite{li2020qeba} is built on top of HSJA~\cite{chen2019hopskipjumpattack} with a dimension-reduced subspace to generate queries for estimating the normal vector direction. QEBA proposes spatial, frequency, and intrinsic component subspaces to better estimate the normal vector. TA~\cite{ma2021tangent} demonstrates a new method for minimizing the $\ell_2$-norm of perturbation by obtaining the tangent of a virtual hemisphere.

\section{Problem definition}
Let a pre-trained $L$-class classifier be modeled as $f(\bm x): \mathbb{R}^n \rightarrow \mathbb{R}^L$. For a given input image $\bm x \in [0,1]^n$, $f \in \mathbb{R}^L$ is the confidence score of the classifier. In a decision-based black-box attack, the classifier only returns the top-1 classification label of $f$. The output of the classifier $f$ for a given query $\bm x$ in the decision-based attack can be expressed as $\hat y (\bm x) = \arg \max_j [f (\bm x)]_j$, where $[f]_j$ is the prediction probability of $j$-th class, $1 \leq j \leq L$. 

For a correctly classified source image $\bm x_s$ by the classifier $f$, the goal is to find a direction $\bm{\hat  \zeta}$ so that $\bm x_s$ can be moved towards that direction to get an adversarial image with minimum perturbation. If a query $\bm x_{q} = \bm x_s + \bm d (\bm {\hat{\zeta}})$ is in the adversarial region, the classifier $f$ returns an incorrect prediction due to the added perturbation, where $\bm d (\bm {\hat{\zeta}})$ denotes the perturbation added to $\bm x_s$.  The optimal direction to get an adversarial image can be formulated as:
\vspace{-1mm}
\begin{equation}
\bm {\hat{\zeta}}^* = \arg\min_{\bm {\hat{\zeta}} \in \mathbb{R}^n} \|\bm d (\bm {\hat{\zeta}})\|_2, \;\;\;\;\; \textrm{s.t.} \; \; \phi(\bm x_{q})=1,
\end{equation}
where $\|\bm d (\bm {\hat{\zeta}})\|_2 = d$ is the $\ell_2$-norm of perturbation added in the direction $\bm {\hat{\zeta}}$, and $\phi(.)$ denotes an indicator function to determine whether the query is correctly classified or misclassified.
For a non-targeted attack: \vspace{-1mm}
\begin{equation}
    \;\; \phi(\bm x_{q})= 
    \begin{cases}
    1, & \text{if  }\; \hat y(\bm x_{q}) \neq \hat y(\bm x_s )\\
    -1, & \text{otherwise}
    \end{cases},
\end{equation}
and for a targeted attack with an intended classification label $l_t$: \vspace{-1mm}
\begin{equation}
    \;\; \phi(\bm x_{q})= 
    \begin{cases}
    1, & \text{if  }\; \hat y(\bm x_{q}) = l_t\\
    -1, & \text{otherwise}
    \end{cases}.
\end{equation}

The optimal direction $\bm {\hat{\zeta}^*}$ results in minimum perturbation $\bm d (\bm {\hat{\zeta}^*})$ to obtain the desired adversarial image $\bm x_{adv}^* = \bm x_s + \bm d (\bm {\hat{\zeta}^*})$. This paper proposes novel methods for obtaining adversarial images for non-targeted and targeted attacks.

\section{Proposed methods}\label{sect:proposed_methods}
Geometric-based attacks like qFool(non-targeted)~\cite{liu2019qFool}, GeoDA~\cite{geoda}, and SurFree~\cite{maho2021surfree} approximate the decision boundary as a hyperplane. However, SurFree doesn't use normal vector information, while qFool and GeoDA lose effectiveness with sufficiently curved boundaries. 
In contrast, CGBA conducts a boundary search along a semicircular path guided by the estimated normal vector, which works effectively for arbitrary decision boundaries, and demonstrates significant improvement on low to medium curvatures. 
Moreover, CGBA is modified to CGBA-H to further adapt to the high curvature setting. Our methods are iterative and estimate the boundary point's normal vector in each iteration, which is one key component accounting for its success.

\paragraph{Normal vector approximation on decision boundary.} \vspace{-2mm}
Let us assume $\bm x_{b_t}$ as a boundary point at the $t$-th iteration. To estimate the normal vector on the decision boundary, we generate $N_t$ number of random samples $\{\bm z_i\}_{i=1}^{N_t}$ from a Gaussian distribution $\bm z_i \sim \mathcal N(0, \mathbf{\sigma^2})$. Then, for each of the samples, we query the classifier with $\bm x_{b_t} + \bm z_i$,  $i \in \{1,..., N_t\}$ to obtain the hard-label prediction of the classifier. Using these queries and their corresponding predictions, the normal unit vector on the boundary at $t$-th iteration can be approximated as~\cite{geoda}:\vspace{-2mm}
\begin{equation} 
    \hat{\bm \eta}_t = \frac{\sum_{i=1}^{N_t} \phi(\bm x_{b_t} + \bm z_i)\bm z_i}{\|\sum_{i=1}^{N_t} \phi(\bm x_{b_t} + \bm z_i)\bm z_i\|_2} .
\end{equation}

\begin{figure}[t] \vspace{0mm}
    \centering
    \includegraphics[trim = 8mm 14mm 4mm 2mm, clip, width=0.3\textwidth]{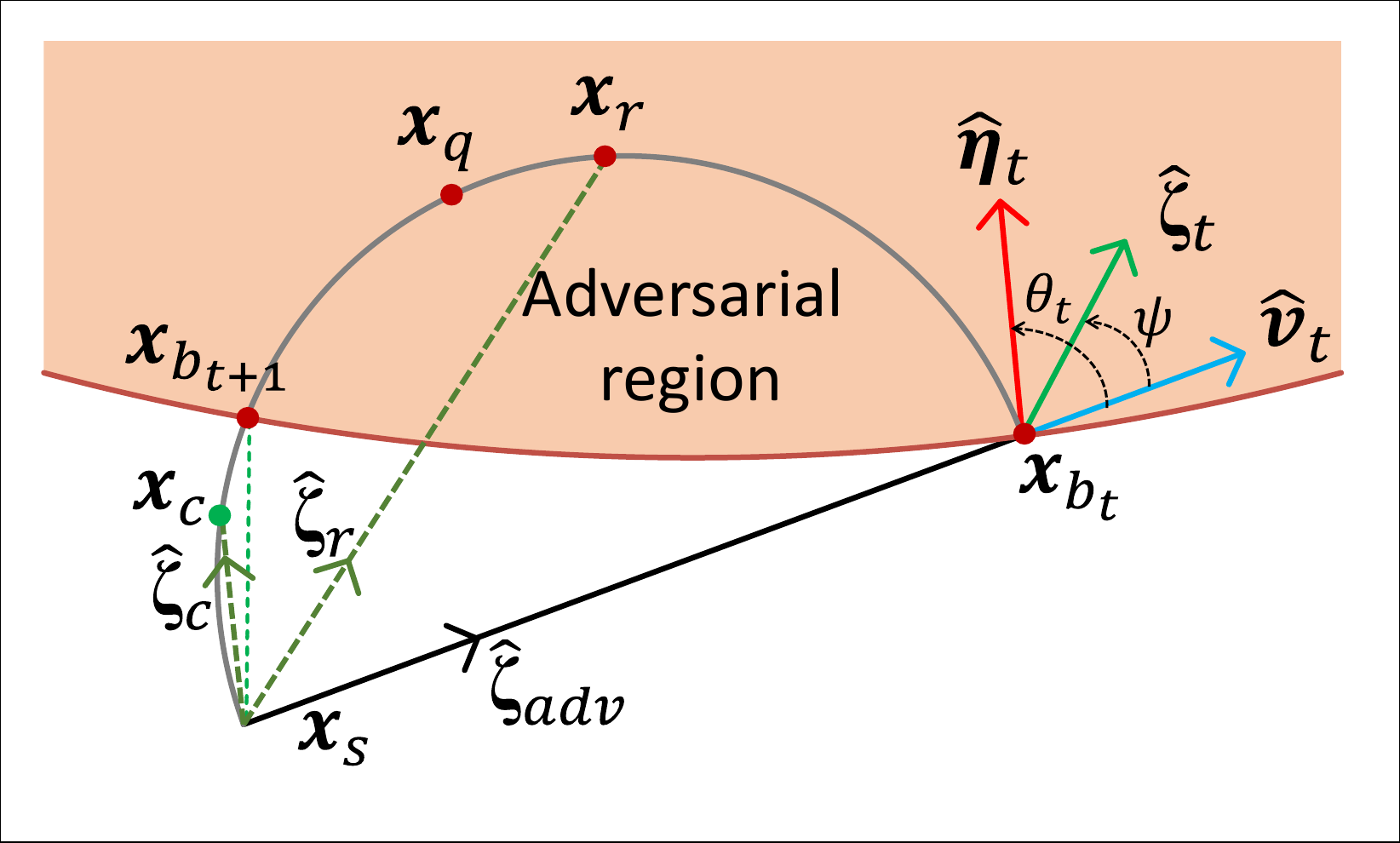} \vspace{-2mm}
    \caption{The geometry of CGBA.}
    \label{al1_mainFig} \vspace{-2mm}
\end{figure}

\subsection{CGBA}

Denote $\hat{\bm v}_t = ({\bm x_{b_t} - \bm x_s})/\|{\bm x_{b_t} - \bm x_s}\|_2$ as the direction of a boundary point $\bm x_{b_t}$ from a source $\bm x_s$, and $\hat{\bm \eta}_t$ is the estimated normal direction on $\bm x_{b_{t}}$ at $t$-th iteration. The key idea of this method is to conduct a boundary search to obtain a better subsequent boundary point $\bm x_{b_{t+1}}$ on a semicircular path in a 2-D plane spanned by $(\hat{\bm \eta}_t, \hat{\bm v}_t)$ on the side of $\hat{\bm \eta}_t$, where the semicircular path is formed between $\bm x_s$ and $\bm x_{b_t}$ centered at $({\bm x_{b_t} + \bm x_s})/2$, as shown in Figure~\ref{al1_mainFig}. 
The search direction to perform a query in the plane spanned by $(\hat{\bm \eta_t},\hat{\bm v_t})$ can be obtained as:\vspace{-2mm}
\begin{equation} 
    \hat{\bm \zeta}_t (m) = \frac{\hat{\bm \eta}_t + m\hat{\bm v}_t}{\|\hat{\bm \eta}_t + m\hat{\bm v}_t\|_2},
\end{equation}
where $m$ is a multiplication factor that controls the search direction. For a search direction $\hat{\bm \zeta}_t$, we can calculate the perturbation $\bm d(\hat{\bm \zeta}_t)$ to obtain the query on the semicircular path as:
\begin{equation} \label{eq:pert_semicircle}
    \bm d(\hat{\bm \zeta_t}) = \| \bm x_{b_t} - \bm x_s\|_2 (\hat{\bm \zeta_t} \cdot \hat{\bm v_t}) \hat{\bm \zeta_t},
\end{equation}
where $\| \bm x_{b_t} - \bm x_s\|_2(\hat{\bm \zeta_t} \cdot \hat{\bm v_t}) = \| \bm x_{b_t} - \bm x_s\|_2 \cos{\psi}$ denotes the added $\ell_2$-norm of perturbation in the direction $\hat{\bm \zeta_t}$ with $\psi$ as the angular difference between $\hat{\bm \zeta}_t$ and $\hat{\bm v}_t$. By varying $\hat{\bm \zeta}_t$, we query the target model for $\bm x_q = \bm x_s + \bm d(\hat{\bm \zeta_t})$ to conduct boundary search on the semicircular path.

The proposed CGBA first finds a non-adversarial point and an adversarial point on the semicircle, and then progressively reduces the range between the adversarial and non-adversarial points to obtain the boundary point $\bm x_{b_{t+1}}$, inspired by the binary search. If $\psi_i$ is the search angle of $\hat{\bm \zeta}_t$ w.r.t. $\hat{\bm v}_t$, the multiplication factor $m_i$ to to attain the search angle $\psi_i$ can be calculated as: \vspace{-1mm}
\begin{equation} 
    m_i = \sin{\theta_t} \cot{\psi_i} - \cos{\theta_t} ; \;\;\; \forall{i} \in \mathbb{Z^+}, \vspace{-2mm}
\end{equation}
where $\theta_t = \cos^{-1}{(\hat{\bm v}_t \cdot \hat{\bm \eta}_t)}$ and $\psi_i = (90^0 -\frac{ 90^0}{2^i}), \forall i \in \mathbb{Z^+}$. With the increase of $i$, the search angle $\psi_i$ is also increased. Thus, for a particular value of $m_i$, we obtain a perturbation $\bm d(\hat{\bm \zeta_t}(m_i))$ and the corresponding point $\bm x_c = \bm x_s + \bm d(\hat{\bm \zeta_t}(m_i))$ on the semicircle such that $\phi(\bm x_c)=-1$.
Then the boundary search between the non-adversarial point $\bm x_c$ and the adversarial point $\bm x_{b_t}$ is conducted by using BSSP to find $x_{b_{t+1}}$ along a semicircular path. 
At the start of the BSSP at $t$-th iteration, consider $\hat{\bm \zeta}_{adv}$ and $\hat{\bm \zeta}_{c}$  are the directions of $\bm x_{b_t}$ and $\bm x_{c}$ from $\bm x_s$, respectively, and $\hat{\bm \zeta}_{r} = (\hat{\bm \zeta}_{adv} + \hat{\bm \zeta}_{c})/\|\hat{\bm \zeta}_{adv} + \hat{\bm \zeta}_{c}\|_2$ is the resultant direction of $\hat{\bm \zeta}_{adv}$ and $\hat{\bm \zeta}_{c}$, as shown in Figure~\ref{al1_mainFig}. 
The BSSP reduces the range of search direction from $[\hat{\bm \zeta}_{adv}, \hat{\bm \zeta}_{c}]$ to $[\hat{\bm \zeta}_{r}, \hat{\bm \zeta}_{c}]$ for $\phi(\bm x_r)=1$ as $\bm x_{b_{t+1}}$ lies between the directions $\hat{\bm \zeta}_{r}$ and $\hat{\bm \zeta}_{c}$, while the range will be reduced to $[\hat{\bm \zeta}_{adv}, \hat{\bm \zeta}_{r}]$ for $\phi(\bm x_r)=-1$, where $\bm x_r = \bm x_s + \bm d (\hat{\bm \zeta}_{r})$. This process of reducing the range of the search direction is continued until obtaining the boundary point $\bm x_{b_{t+1}}$ with a certain accuracy.
One important characteristic of BSSP is that it ensures $\bm x_{b_{t+1}}$ with a reduced perturbation since for any query $\bm x_q$ on the semicircular path, $\|\bm x_q - \bm x_s\|_2 \leq \|\bm x_{b_t} - \bm x_s\|_2$.

\begin{algorithm}[t]
\small	
	\DontPrintSemicolon
	 \textbf{Inputs:} Source image $\bm x_s$, indicator function $\phi(.)$, a random direction $\bm \Theta$, queries to estimate initial normal vector $N_{0}$, iteration $T$.
	 
	 \textbf{Output:} Adversarial example $\bm x_{adv}$.

	
	 $r \gets \min \{r>0: \phi(\bm x_s + r*\frac{\bm \Theta}{\|\bm \Theta \|_2}) =1\}$\\
	 $\bm x_{b_1} = \bm x_s + r*\frac{\bm \Theta_{1}}{\|\bm \Theta_{1} \|_2}$\\
	\For{$t=1:T$}{
	    Generate $N_t = N_{0} \sqrt{t}$ samples,  $\bm{z_i} \sim \mathcal{N}(0, \mathbf{\sigma^2})$
	    
		Estimate $\hat {\bm \eta}_{t}$ using  $\bm{z_i}$ at $\bm x_{b_{t}}$ by $N_t$ queries.\;
		$\hat{\bm v}_t = \frac{\bm x_{b_t} - \bm x_s}{\| \bm x_{b_t} - \bm x_s \|_2}$, ~ $\theta_t = \cos^{-1}(\hat{\bm \eta}_t \cdot \hat{\bm v}_t)$, ~ $i =1$
		
        \While{True}{
        $m_i = \sin{\theta_t} \cot{(90^0 - \frac{90^0}{2^i})} - \cos{\theta_t}$\;
        $ \hat{\bm \zeta}_{t} = (\hat {\bm \eta}_{t} + m_i  \hat{\bm v}_t)/\|\hat {\bm \eta}_{t} + m_i  \hat{\bm v}_t\|_2$\;
        $\bm x_{q} = \bm x_s + \bm d(\hat{\bm \zeta}_t)$, \quad $i=i+1$\;
          \If{$\phi(\bm x_{q}) = -1$}{
            break
          }
        }
			
		$\bm x_{b_{t+1}} \gets BSSP(\bm x_s, \bm x_{q}, \bm x_{b_t}, \phi)$ \Comment*[r]{to find the boundary point on semicircular path} 
		
		}
	$\bm x_{adv} = \bm x_{b_{t+1}}$

	\caption{CGBA}
	\label{alg:1}
\end{algorithm}

\begin{figure}  \vspace{-2mm}
    \centering
    \subfloat[]{\includegraphics[trim = 2mm 6mm 5mm 2mm, clip,width=0.22\textwidth]{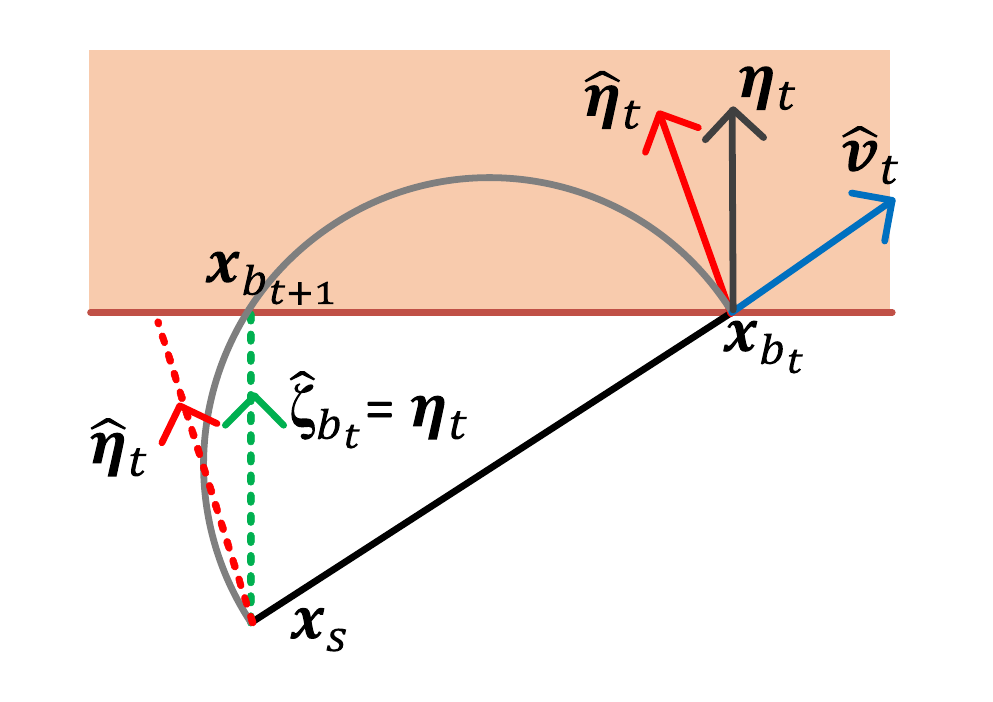}\vspace{-2mm}\label{fig:lin_hplane}}
    \subfloat[]{\includegraphics[trim = 20mm 2mm 25mm 20mm, clip,width=0.2\textwidth]{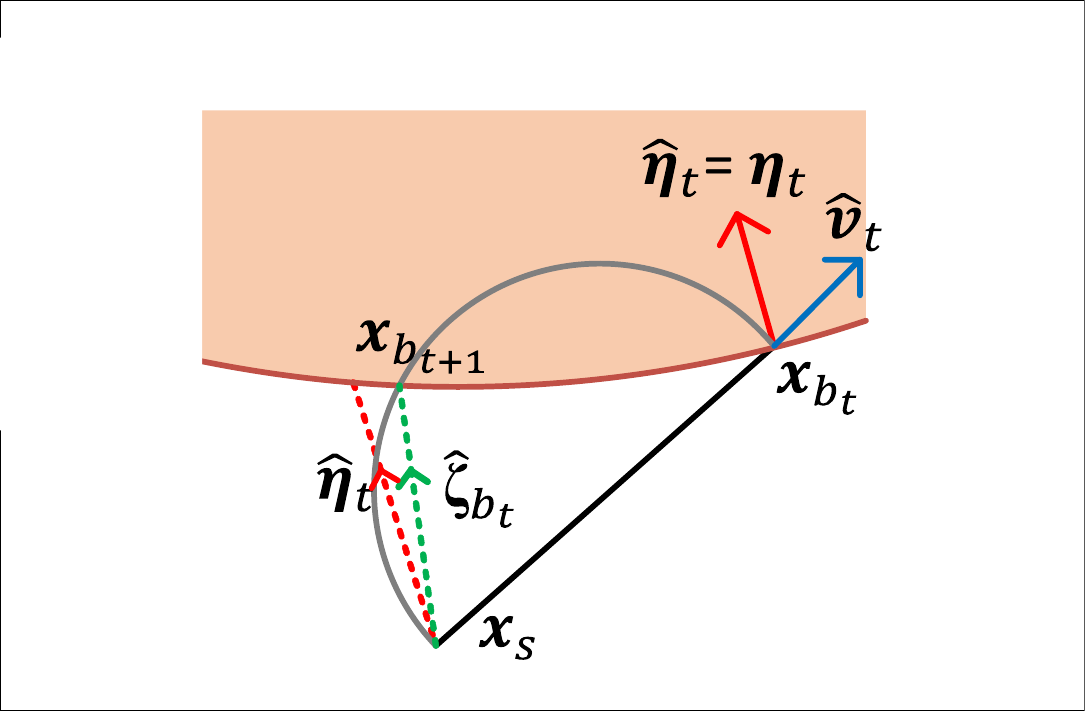} \label{fig:curved_hplane}} \vspace{-2mm}
    \caption{(a) Linear and (b) low curvature boundaries.}
    \label{fig:Lin_lowCurv_boundary} \vspace{-2mm}
\end{figure}


To demonstrate the efficacy of the BSSP algorithm, we consider two scenarios of boundary: linear and low curvature boundaries. 
Let us consider $\hat {\bm \zeta}_{b_{t}}$ as the direction of $\bm x_{b_{t+1}}$ obtained by using the BSSP method. Thus, $\bm x_{b_{t+1}}$ can be calculated as: $ \bm x_{b_{t+1}} = \bm x_s + \bm d(\hat {\bm \zeta}_{b_{t}})$.  
First of all, in the case of a linear boundary, the direction $\hat{\bm \zeta}_{b_{t}}$ makes a right angle with the boundary as any inscribed angle in a semicircle makes a right angle, as shown in Figure~\ref{fig:lin_hplane}. If $\bm \eta_t$ represents the true normal vector on the boundary, then the direction of  $\bm \eta_t$ coincides with $\hat {\bm \zeta}_{b_{t}}$. 
Thus, it should be enough to push $\bm x_s$ towards $\bm \eta_t$ using the BSNV method to get the optimal perturbation as it is done in~GeoDA\cite{geoda} and qFool (non-targeted)\cite{liu2019qFool}. 
However, $\bm \eta_t$ is an unknown parameter that can be approximated using the normal vector estimation process. There is a high chance that the estimated normal vector direction $\hat{\bm \eta_t}$ deviates from $\bm \eta_t$. 
From Figure~\ref{fig:lin_hplane}, if there is a deviation between $\bm \eta_t$ and $\hat {\bm \eta_t}$, pushing $\bm x_s$ towards $\hat {\bm \eta_t}$ will not result in the optimal $\bm x_{b_{t+1}}$. In contrast, querying on the semicircular path in the plane spanned by $(\hat{\bm v_t}, \hat{\bm \eta_t})$ finds optimal $\bm x_{b_{t+1}}$ in that plane.
Secondly, if we consider a low curvature decision boundary, as shown in Figure~\ref{fig:curved_hplane}, BSSP finds a boundary point with smaller perturbation than using binary search towards $\hat{\bm 
\eta}_t$  even $\hat{\bm 
\eta}_t$ is same as the $\bm {\eta}_t$. 


Considering the above scenarios, the BSSP finds a better boundary point than the BSNV, which in turn makes the proposed CGBA effective. 
Experimental and theoretical evidence supporting that BSSP is more effective than BSNV are provided in Appendix~\ref{bsssp_bsnv}. 
The pseudocode of CGBA for the non-targeted attack is given in Algorithm~\ref{alg:1} which can be easily converted to the targeted attack. The pseudocode of the BSSP algorithm is given in Appendix~\ref{algorithms}.

\begin{figure}[h] \vspace{-2mm}
    \centering
    \subfloat[]{\includegraphics[trim = 20mm 45mm 30mm 30mm, clip,width=0.22\textwidth]{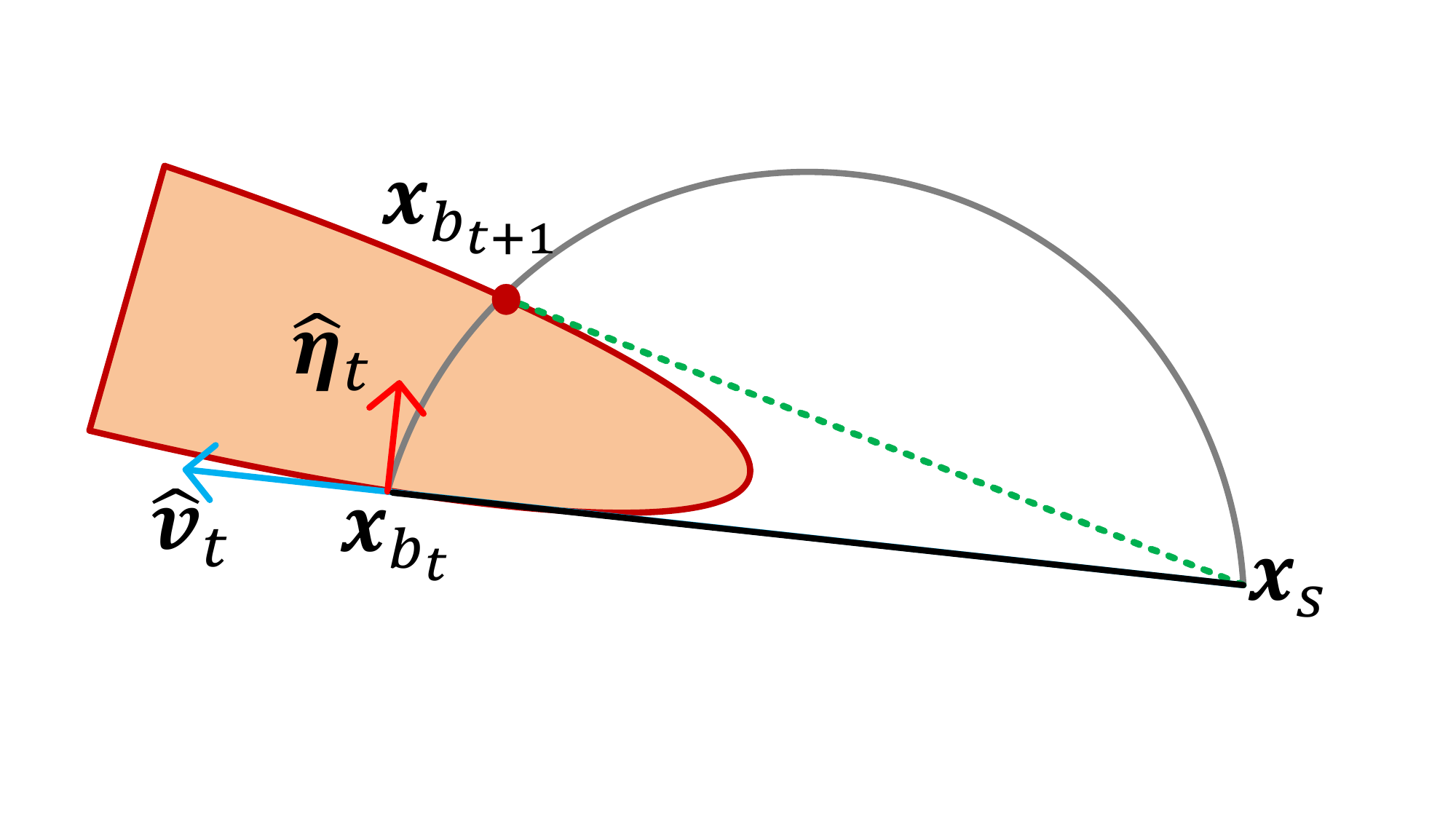}\label{fig:high_curve1}}
    \subfloat[]{\includegraphics[trim = 20mm 50mm 35mm 30mm, clip,width=0.22\textwidth]{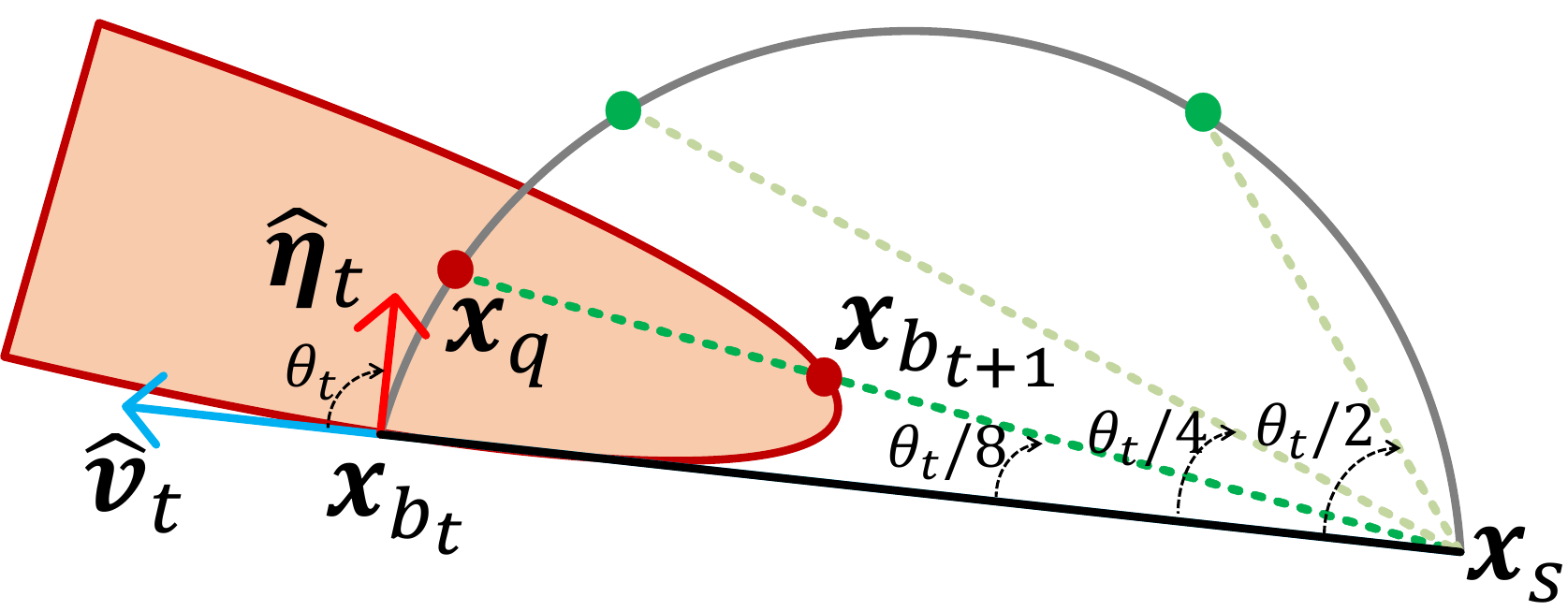}\label{fig:high_curve2}}
    \caption{Boundary with high curvature.}
    \label{fig:method2} \vspace{-2mm}
\end{figure}


\subsection{CGBA-H} 
Adversarial attacks using CGBA can be an effective approach for the decision boundary with low curvature. However, this approach becomes less effective if the curvature of the boundary is too high. 
From Figure~\ref{fig:high_curve1}, it can be realized that obtaining the boundary point $\bm x_{b_{t+1}}$ using the BSSP may result in a boundary point that is away from the optimal solution. 
To avoid this situation, we propose a more effective approach to get a better boundary point in each iteration for the boundary with high curvature. 
If $\theta_t = \cos^{-1}(\hat{\bm \eta}_t \cdot \hat{\bm v}_t)$ is the angle between $\hat{\bm \eta}_t$ and $\hat{\bm v}_t$, then the multiplying factor to estimate the direction to query can be calculated as: \vspace{-1mm}
\begin{equation} \vspace{-2mm}
    m_i = \sin{\theta_t} \cot{\left\{\frac{\theta_t}{2^i}\right\}} - \cos{\theta_t}; \;\;\; \forall{i} \in \mathbb{Z^+},
\end{equation}
where the value of $i$ ensures $\cos^{-1}(\hat {\bm v}_t \cdot \hat{\bm \zeta}_t(m_i)) = \theta_t / {2^i}; \; \forall{i} \in \mathbb{Z^+}$. So, with the increase of $i$, the angular difference between $\hat {\bm v}_t$ and estimated $\hat{\bm \zeta}_t(m_i)$ can be reduced. 
Thus, in each iteration, the proposed CGBA-H finds a multiplication factor $m_i$ and the corresponding $\bm x_q = \bm x_s + \bm d \hat{\bm \zeta}_t(m_i)$ on the semicircular trajectory such that $\phi (\bm x_q)=1$, as shown in Figure~\ref{fig:high_curve2}. Then, by conducting the binary search between $\bm x_s$ and $\bm x_q$, a better boundary point $\bm x_{b_{t+1}}$ can be obtained than CGBA.
The pseudocode of CGBA-H for the targeted attack is given in Algorithm~\ref{alg:method2}.

\subsection{Initialization} \label{init}
The initial boundary point $\bm x_{b_1}$ may have a significant impact on the performance of an adversarial attack. In the existing normal vector-based targeted attack, a binary search in the direction of a randomly chosen image of the target class from the source image $\bm x_s$ is conducted to obtain initial boundary point $\bm x_{b_1}$. Rather than finding $\bm x_{b_1}$ in the direction of a randomly chosen target sample form $\bm x_s$, a set of $K$ random directions $\{\bm \Theta_k\}_{k=1}^K$ towards the adversarial region by using $K$-number of samples of the target class can be used to find the direction that provides a boundary point with reduced perturbation for the targeted attack. 
Experimental results reveal that just using a few samples of the target class to obtain $\bm x_{b_1}$ can significantly improve the performance of a decision-based adversarial attack. 
The pseudocode of the Initialization method to obtain the first boundary point is given in Appendix~\ref{algorithms}.

\section{Experiments} \label{sec:experiments}
In this section, we perform a comprehensive set of experiments and compare the results with state-of-the-art algorithms to demonstrate the effectiveness of our proposed methods for non-targeted and targeted attacks. Moreover, we show how initialization affects the performance of CGBA and CGBA-H.

\begin{table*}[]
\scriptsize
    \centering
\begin{tabular}{c|c|c c c c c c c|c c c c c c c|}
    \toprule
      \multicolumn{2}{r|}{Attack}  & \multicolumn{7}{c|}{Non-targeted} & \multicolumn{7}{c|}{Targeted}\\
      \hline
    \multicolumn{2}{r|}{Queries}  & 1000 & 2500 & 5000 & 7500 & 10000 & 15000 & 20000 & 1000 & 2500 & 5000 & 7500 & 10000 &15000 & 20000\\
    \midrule \midrule
    \multirow{7}{*}{\rotatebox[origin=c]{90}{ResNet50}} 
     
    & HSJA~\cite{chen2019hopskipjumpattack}  & 13.42 & 6.46 & 3.76 & 2.93 & 2.49 & 2.04 & 1.79 & 64.27 & 51.54 & 34.58 & 24.51 & 17.68 & 11.15 & 7.99\\ 
    & GeoDA~\cite{geoda}  & 8.41 & 4.72 & 3.54 & 2.93 & 2.71 & 2.39 & 2.20 & - & - & - & - & - & - & -\\\
    & TA~\cite{ma2021tangent}  & 13.98 & 6.36 & 3.77 & 2.97 & 2.46 & 1.97 & 1.70 & 63.09 & 46.55 & 31.94 & 23.05 & 16.95 & 10.91 & 7.87\\ 
    & TriA~\cite{wang2022Triangle}  & 6.26 & 5.58 & 5.39 & 5.15 & 5.03 & 4.84 & 4.73 & - & - & - & - & - & - & -\\
    & SurFree~\cite{maho2021surfree}   & 8.44 & 4.42 & 2.65 & 1.96 & 1.58 & 1.17 & 0.97 & - & - & - & - & - & - & -\\
    & AHA~\cite{li2021aha} & - & - & - & - & - & - & - & 56.55 & 37.91 & 23.04 & 15.48 & 11.46 & 8.76 & 8.23\\ 
    & CGBA   & 6.03 & \textbf{2.55} & \textbf{1.44} & \textbf{1.05} & \textbf{0.86} & \textbf{0.68} & \textbf{0.59} & 78.99 & 63.60 & 41.71 & 26.04 & 17.26 & 8.72 & 5.38\\
    & CGBA-H   & \textbf{5.78} & 2.67 & 1.51 & 1.11 & 0.91 & 0.73 & 0.62& \textbf{56.01} & \textbf{36.86} & \textbf{21.83} & \textbf{13.71} & \textbf{9.47} & \textbf{5.63} & \textbf{4.03}\\

    \midrule 
    \multirow{7}{*}{\rotatebox[origin=c]{90}{VGG16}} & HSJA\cite{chen2019hopskipjumpattack} & 8.58 & 4.11 & 2.54 & 2.06 & 1.75 & 1.44 & 1.29 & 65.10 & 47.31 & 30.61 & 21.72 & 15.58 & 9.72 & 7.13\\
    & GeoDA~\cite{geoda} & 5.74 & 3.41 & 2.49 & 2.11 & 1.98 & 1.67 & 1.63 & - & - & - & - & - & - & -\\
    & TA~\cite{ma2021tangent} & 8.44 & 4.09 & 2.57 & 2.07 & 1.77 & 1.45 & 1.30 & 61.97 & 44.42 & 28.62 & 19.41 & 15.03 & 9.70 & 7.13\\ 
    & TriA~\cite{wang2022Triangle}  & 7.42 & 5.54 & 4.95 & 4.63 & 4.41 & 4.30 & 4.20 & - & - & - & - & - & - & -\\
    & SurFree~\cite{maho2021surfree}  & 6.01 & 3.18 & 1.96 & 1.52 & 1.24 & 0.97 & 0.82 & - & - & - & - & - & - & -\\
    & AHA~\cite{li2021aha} & - & - & - & - & - & - & - & 55.40 & 36.46 & 21.37 & 14.26 & 10.91 & 8.73 & 8.34\\
    & CGBA & 3.99 & \textbf{1.86} & \textbf{1.08} & \textbf{0.82} & \textbf{0.69} & \textbf{0.57} & \textbf{0.50} & 80.13 & 67.09 & 44.93 & 27.67 & 16.33 & 7.69 & 4.91\\
    & CGBA-H & \textbf{3.93} & 1.94 & 1.17 & 0.89 & 0.75 & 0.61 & 0.54 & \textbf{52.82} & \textbf{33.29} & \textbf{18.09} & \textbf{11.22} & \textbf{7.79} & \textbf{4.92} & \textbf{3.67}\\

     \bottomrule 
\end{tabular}   
    \caption{Median $\ell_2$-norm of perturbation for different query budgets against ResNet50 and VGG16 on ImageNet dataset.}
    \label{tab:1_norm_budget} \vspace{-1mm}
\end{table*}

\subsection{Experimental setting} \vspace{-4mm}
\noindent \paragraph{Datasets and target models.} We evaluate the performance of CGBA and CGBA-H using ImageNet~\cite{deng2009imagenet} and CIFAR-10~\cite{krizhevsky2009learning} datasets. 
The performance of the proposed attacks on the ImageNet dataset is evaluated using pretrained ResNet50~\cite{he2016deep}, VGG16~\cite{vgg16_simonyan2014very}, ResNet101~\cite{he2016deep} and ViT~\cite{ViT_dosovitskiy2020image} classifiers. 
The first three pretrained classifiers can be found in the PyTorch, and ViT is obtained from the PyTorch Image Models\footnote{\url{https://github.com/rwightman/pytorch-image-models}}. 
For each target model, we randomly select 1000 images for the non-targeted attack and 1000 pairs of images for the targeted attack from the ILSVRC2012's validation set~\cite{deng2009imagenet} that are correctly classified by the target model.
The images are resized to 3 $\times 224 \times 224$ as an input to the classifiers. 
For the CIFAR-10 dataset, we consider PreActResNet-18~\cite{he2016identity} and a wide residual network with 40 layers (WRN40)~\cite{zagoruyko2016WRN} as target classifiers. We train both classifiers for 200 epochs with an image resolution of $3 \times 32 \times 32$. The proposed attacks on CIFAR10 are also evaluated using a randomly chosen 1000 correctly classified images for the non-targeted attack and 1000 pairs of correctly classified images for the targeted attack.

\begin{algorithm}[t]
\small	
	\DontPrintSemicolon
	 \textbf{Inputs:} Source image $\bm x_s$, a random image $\bm x_t$ of target class $l_t$, indicator function $\phi(.)$,  queries to find initial normal vector $N_{0}$, iteration $T$.
	 
	 \textbf{Output:} Adversarial example $\bm x_{adv}$.


	 $\bm x_{b_{1}} \gets BinarySearch (\bm x_s, \bm x_t, \phi)$ \Comment*[r]{to find initial boundary point} 
	
	\For{$t=1:T$}{
	    Generate $N_t = N_{0} \sqrt{t}$ samples,  $\mathbf{z_i} \sim \mathcal{N}(0, \mathbf{\sigma^2})$
	    
		Estimate $\hat {\bm \eta}_{t}$ using $\mathbf{z_i}$ at $\bm x_{b_{t}}$ by $N_t$ queries.\;
		$\hat {\bm v}_t = \frac{\bm x_{b_{t}} - \bm{x_s}}{\| \bm x_{b_{t}} - \bm{x_s}\|_2}$, ~ $\theta_t = \cos^{-1} (\hat{\bm \eta_t}\cdot \hat {\bm v}_t )$, ~ $i=1$\;

        \While{True}{
        $m_i =\sin{\theta_t} \cot \left( \frac{\theta_t}{2^{i}} \right) - \cos{\theta_t} $\;
        $ \hat{\bm \zeta}_t = (\hat {\bm \eta}_{t} + m_i \hat {\bm v}_t)/ \|\hat {\bm \eta}_{t} + m_i \hat {\bm v}_t\|_2$\;
        $\bm x_{q} = \bm x_s + \bm d(\hat{\bm \zeta}_{t} )$, ~ $i = i+1$\;
          \If{$\phi(\bm x_{q}) = 1$}{
            break
          }
        }
			
		$\bm x_{b_{t+1}} \gets BinarySearch(\bm x_s, \bm x_{q}, \phi)$ \Comment*[r]{to find boundary point} 
		
	} 
	$\bm x_{adv} = \bm x_{b_{t+1}}$
	\caption{CGBA-H}
	\label{alg:method2} 
\end{algorithm}

\noindent \paragraph{Baselines and hyper-parameter setting.} 
We compare the performance of CGBA and CGBA-H with the existing state-of-the-art non-targeted and targeted attacks. We choose HSJA~\cite{chen2019hopskipjumpattack}, GeoDA~\cite{geoda},  generalized TA ~\cite{ma2021tangent}, TriA~\cite{wang2022Triangle}, SurFree~\cite{maho2021surfree} and AHA~\cite{li2021aha} as baselines to compare. 
Among the baselines, HSJA and TA are available for both non-targeted and targeted attacks. However, GeoDA, TriA and SurFree are only given for the non-targeted attack~\cite{ma2021tangent}, while AHA is available for the targeted attack. 
We consider GeoDA, SurFree and AHA for dimension-reduced subspace as these algorithms are given for this setting. For an image with a dimension of $3 \times 224 \times 224$, the reduced dimension by a factor $f$ is given as $3 \times \frac{224}{f} \times \frac{224}{f}$.
GeoDA and SurFree use dimension-reduced frequency subspace by reducing the dimension with a factor $f=5.17$ and $f=2$ to obtain coefficients of DCT transform, respectively, as their default setting. In contrast, AHA reduces the dimension in the spatial subspace by a factor $f=4$ as their best setting.  
For our proposed attacks, we reduce the dimension by $f=4$ in frequency subspace. 
We also set queries to estimate the initial normal vector as $N_0=30$ and the standard deviation for generating random samples from the Gaussian distribution as $\sigma=0.0002$ to estimate the normal vector. 



We use three metrics—median $\ell_2$-norm of perturbation, attack success rate (ASR), and area under the curve (AUC)—to evaluate the performance of CGBA and CGBA-H with SOTA black-box attacks. The median of the $\ell_2$-norm of perturbation for a given query budget using an attack determines the effectiveness of the attack. An attack with better capability to reduce the $\ell_2$-norm of perturbation on a set of test images is deemed as a more effective attack. In addition, another popular metric, ASR, is used to determine the success rate of an adversarial attack for a given query budget and perturbation threshold. An attack is considered successful if the obtained perturbation for a particular query budget falls below the perturbation threshold.  Moreover, AUC—the area under the curve of the median $\ell_2$-norm of perturbation versus queries—demonstrates the convergence toward minimum perturbation of an attack with the number of queries. The lower the value of an attack's AUC, the faster the attack converges to the minimum perturbation.

\begin{table*} 
\footnotesize
\centering
    \begin{tabular}{c|c|c  @{\hspace{1.15\tabcolsep}} c @{\hspace{1.15\tabcolsep}} c @{\hspace{1.15\tabcolsep}} c @{\hspace{1.15\tabcolsep}} c @{\hspace{1.15\tabcolsep}} c @{\hspace{1.15\tabcolsep}} c @{\hspace{1.35\tabcolsep}} c|}
    \toprule
        &Methods & HSJA~\cite{chen2019hopskipjumpattack} & GeoDA~\cite{geoda} & TA~\cite{ma2021tangent} & TriA~\cite{wang2022Triangle} & SurFree~\cite{maho2021surfree} & AHA~\cite{li2021aha} & CGBA & CGBA-H  \\ 
        \midrule
        \multirow{2}{*}{{ResNet50}} & Non-targeted & 86531 & 129362 &  86756 & 107080 & 54575  & -  &  37195 &  \textbf{36460} \\
        & Targeted & 520616  & -  &  486946 & - & -  &  383985 &  558611 &  \textbf{341965}  \\
        \hline
        \multirow{2}{*}{{VGG16}} & Non-targeted & 59049 & 54545 & 58987 & 98051 & 41270 & - & \textbf{27275} & 27604 \\
        & Targeted & 471494 & - & 451480 & - & - & 370967 &  566433 &  \textbf{304521} \\
        \hline
        \multirow{2}{*}{{ResNet101}} & Non-targeted & 96710 & 79915 & 97221 & 115203 & 66382 & - & 46229 & \textbf{43250}  \\
        & Targeted & 568382 & - & 504363 & - & - & 399841 &  541873 &  \textbf{353937} \\
        \hline
        \multirow{2}{*}{{ViT}} & Non-targeted & 152219 & 130957 & 156360 & 129956 & 94329 & - & 63148  & \textbf{55372} \\
        & Targeted & 447887 & - & 394218 & - & - & 298171 & 368239 &  \textbf{283216} \\
        \bottomrule
    \end{tabular} \vspace{-2mm}
    \caption{AUC comparison against ResNet50, VGG16, ResNet101 and ViT for a query budget of 20000 on ImageNet.}
    \label{tab:auc_resnet50}
\end{table*}

\begin{figure*} \vspace{-3mm}
    \centering
    \subfloat[Non-targeted attack]{\includegraphics[width=0.235\textwidth]{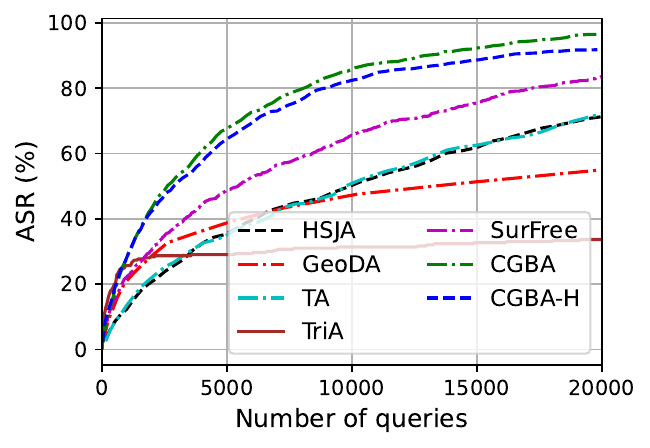}}
    \hfill
    \subfloat[Targeted attack]{\includegraphics[width=0.235\textwidth]{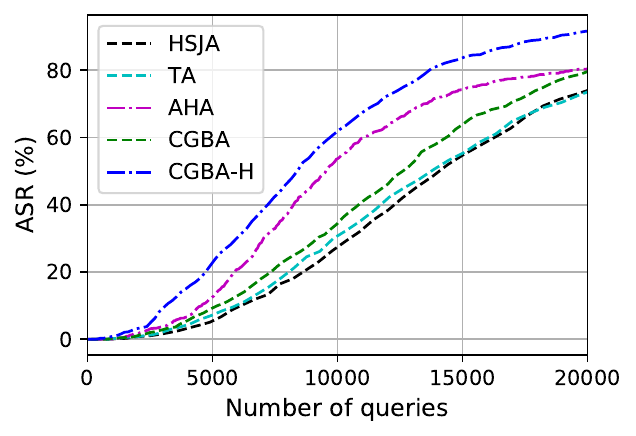}}
    \hfill
    \subfloat[Non-targeted attack]{\includegraphics[width=0.235\textwidth]{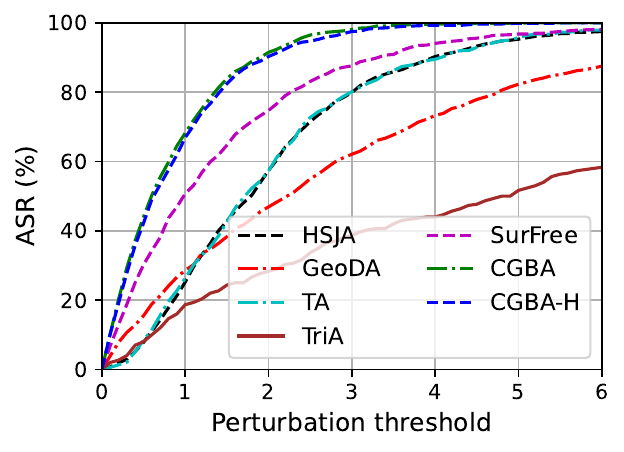}}
    \hfill
    \subfloat[Targeted attack]{\includegraphics[width=0.235\textwidth]{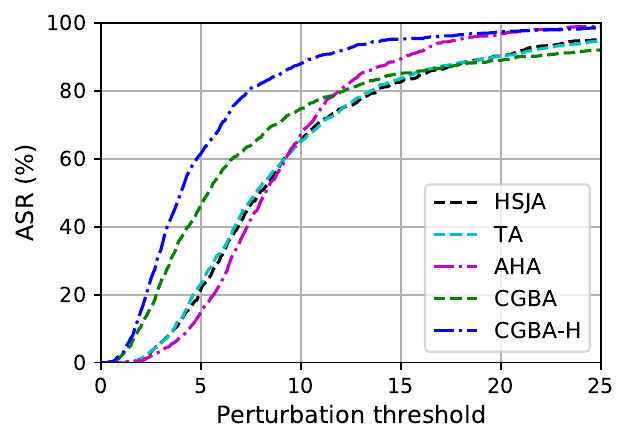}} \vspace{-2mm}
    \caption{ASR versus queries (a-b), and ASR versus perturbation thresholds (c-d) against ResNet50 on ImageNet.}
    \label{fig:resnet50_asr}
\end{figure*}

\subsection{Experimental results}
 Table~\ref{tab:1_norm_budget} presents the median $\ell_2$-norm of perturbation for different query budgets obtained by various baselines and our proposed algorithms for both non-targeted and targeted attacks, evaluated against ResNet50 and VGG16 models using the ImageNet dataset. 
 For further information, the median $\ell_2$-norm of perturbation against ResNet101 and ViT models, along with the corresponding curves for all classifiers, can be found in Appendix~\ref{imagenet_results}. Additionally, Appendix~\ref{cifar_results} contains the experimental results on CIFAR10.
 
 In the case of non-targeted attacks, we observe that CGBA and CGBA-H outperform all the baselines.
 In most cases, with a sufficient query budget, CGBA achieves the best performance. When the query budget is limited, CGBA-H offers better performance; intuitively a lower-quality boundary point is obtained with a limited budget and the boundary appears more curved from the viewpoint of the source image. For targeted attacks, CGBA-H achieves the best performance across the board, and the gap with the baselines increases with the increase of query budget. It is interesting to note that even CGBA outperforms the baselines of targeted attacks with a sufficiently large query budget; intuitively the boundary appears much flatter from the viewpoint of the source image in this case with high-quality boundary points. The above observations conform to our insights and verify the effectiveness of our proposed methods.

\begin{figure} \vspace{-3mm}
    \centering
    \subfloat[Gray-whale misclassified as an arbitrary class Stole.]{\includegraphics[width=0.48\textwidth]{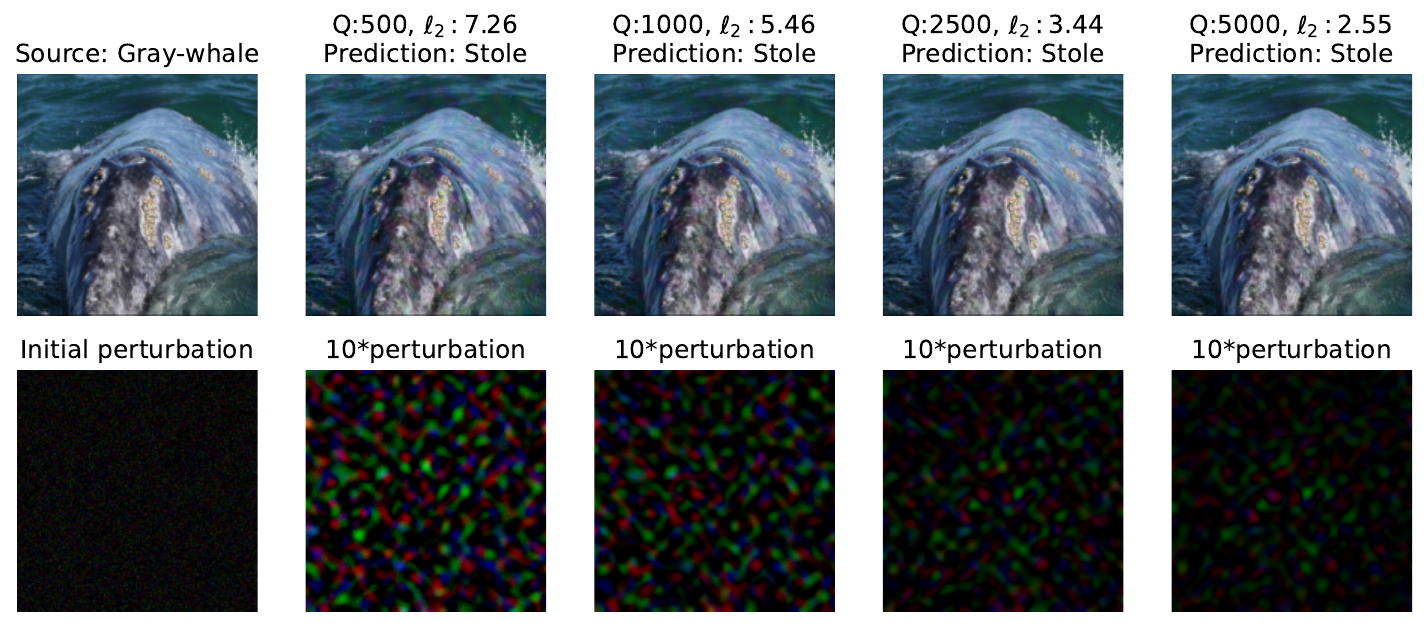}\label{fig:adv_nonTar}} 
    \vfill \vspace{-3mm}
    \subfloat[Spoonbill misclassified as target class Bee-eater.]{\includegraphics[width=0.48\textwidth]{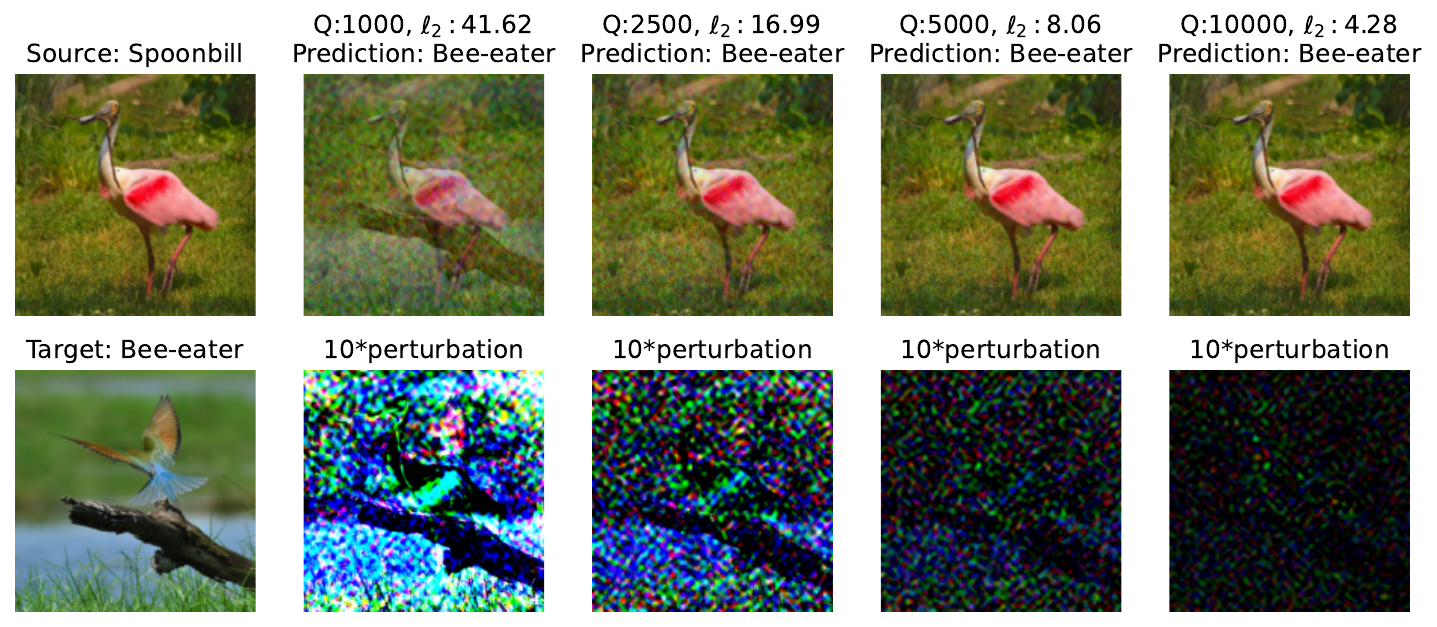}\label{fig:adv_Tar}} \vspace{-3mm}
    \caption{Adversarial examples for different query budgets.}
    \label{fig:adv_samples} 
    \vspace{-5mm}
\end{figure}

Figure~\ref{fig:resnet50_asr} demonstrates the ASR comparison of the proposed methods with the baselines against ResNet50, and Appendix~\ref{imagenet_results} contains corresponding curves for the other classifiers on ImageNet. The left two sub-figures show the impact of queries on ASR. To plot these figures, we consider a perturbation threshold of 2.5 for non-targeted attacks and 12 for targeted attacks. The right two sub-figures, on the other hand, show the variation of ASR with different threshold values for a query budget of 20,000. These figures further demonstrate the superiority of CGBA and CGBA-H for non-targeted attacks and targeted attacks, respectively. 
Similar observations can be made from Table~\ref{tab:auc_resnet50} on AUC comparison.  
{Furthermore, we obtained perturbed images by using non-targeted CGBA and targeted CGBA-H for different query budgets, as shown in Figure~\ref{fig:adv_nonTar} and \ref{fig:adv_Tar}. In Figure~\ref{fig:adv_samples},  $Q$ denotes a query budget, and $\ell_2$ denotes the amount of perturbation corresponding to that query. We depict amplified obtained perturbation to observe how the perturbation diminishes with the increase of query budgets starting from an arbitrary random noise for the non-targeted attack and starting from a target image for the targeted attack.   }   

\begin{figure}[t] \vspace{-5mm}
    \centering
    \subfloat[Non-targeted attack]{\includegraphics[width=0.23\textwidth]{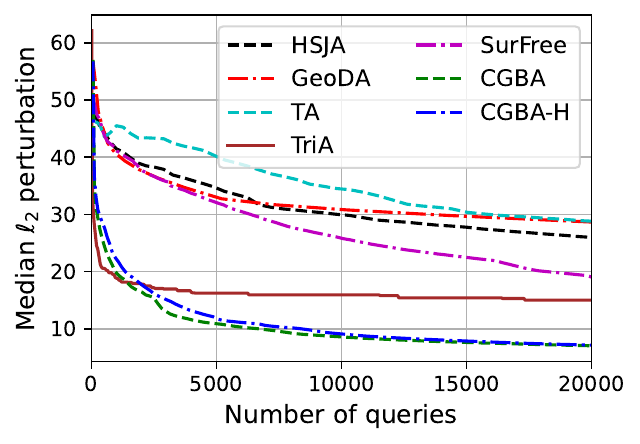}\label{fig:AT:1}}
    \hfill
    \subfloat[Targeted attack]{\includegraphics[width=0.23\textwidth]{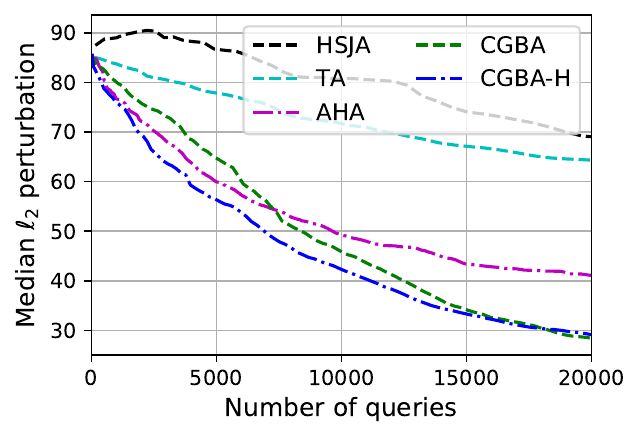}\label{fig:AT:2}} \vspace{-2mm}
    \caption{Results against an adversarially-trained model.}
    \label{fig:AT} \vspace{-5mm}
\end{figure}

\vspace{-2mm}
\paragraph{Performance against adversarially-trained model.}
 One of the most popular defense methods against adversarial attacks is adversarial training~\cite{madry2018towards}. To evaluate the performance of the proposed attacks against the adversarially-trained model, we used a pre-trained ResNet50 model from the GitHub repository of MardyLab\footnote{\url{https://github.com/MadryLab/robustness}}. We randomly choose 200 samples for the non-targeted attack and 200 pairs of samples for the targeted attack. Figure~\ref{fig:AT:1} shows that our proposed methods perform much better than the SOTA baselines. One plausible explanation is that adversarial training makes the boundary flatter~\cite{moosavi2019robustnesssmoothness}, and CGBA, guided by the normal vector, makes the best use of the flatness of the boundary. From Figure~\ref{fig:AT:2}, it is observed that the proposed methods are also effective in performing targeted attacks against the adversarially-trained model. 

\begin{figure} \vspace{-2mm}
    \centering
    \subfloat{\includegraphics[trim = 10mm 5mm 0mm 5mm, clip,width=0.49\textwidth]{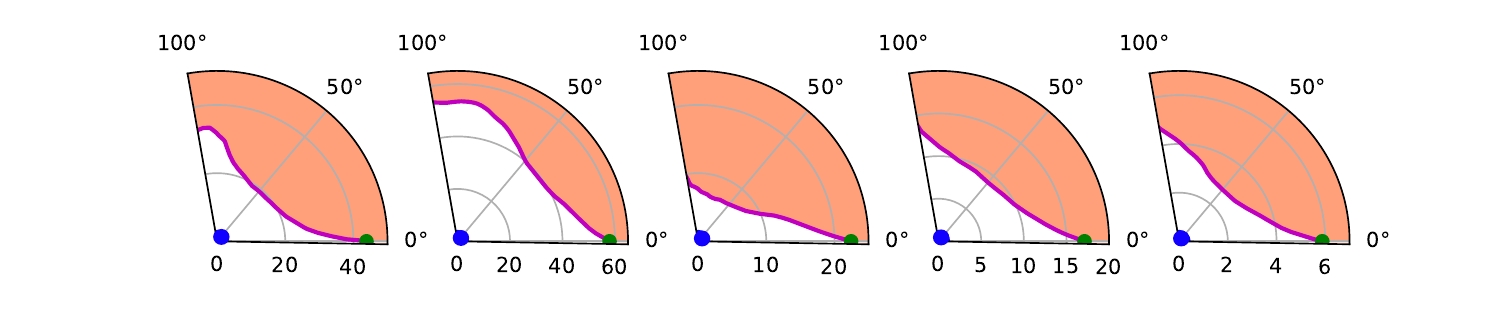}\label{fig:non_targeted_boundary}} \vfill \vspace{-2mm}
    \subfloat{\includegraphics[trim = 10mm 5mm 0mm 5mm, clip,width=0.49\textwidth]{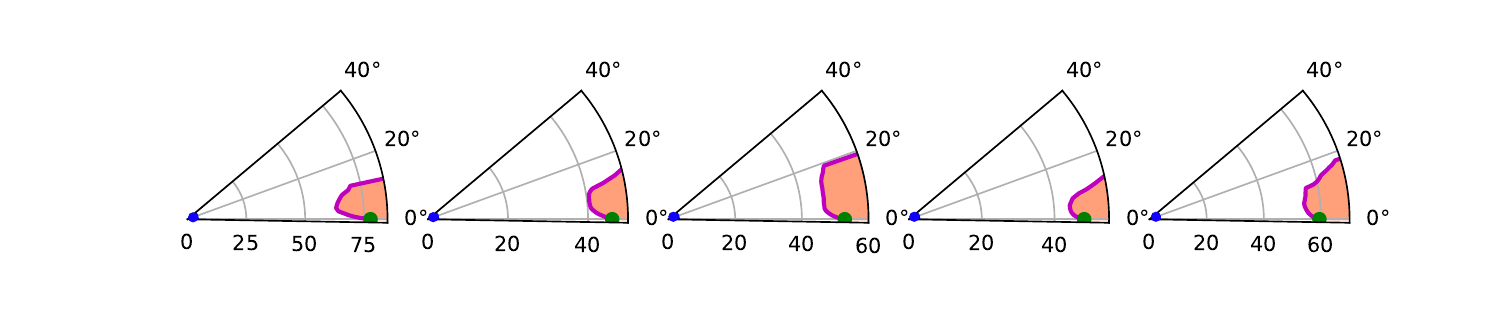} \label{fig:targeted_boundary}} \vspace{-3mm}
    \caption{Boundary trajectory for non-targeted (top-row) and targeted (bottom-row) attacks.}
    \label{fig:boundary}
\end{figure}

 \vspace{-1mm}
\paragraph{Boundary trajectory.} \vspace{-2mm}
We demonstrate the difference in boundary trajectory between non-targeted and targeted attacks. We pick a set of five images and another set of five-pair of images randomly for non-targeted and targeted attacks, respectively. As the proposed methods are based on finding the normal vector on the decision boundary, on observing the boundary trajectory for a single iteration, we consider 400 queries to estimate  $\bm{\hat{\eta}}_1$. In Figure~\ref{fig:boundary}, the blue dotted point at the center and green dotted point in each of the sub-figures indicate the source $\bm x_s$ and the initial boundary point $\bm x_{b_1}$, respectively. For a particular image, the direction of the green point from the blue point is denoted by $\bm{\hat{v}}_1$, and we consider this direction as a reference direction with 0-degree. After obtaining $\bm{\hat{\eta}}_1$ and $\bm{\hat{v}}_1$, we conduct a search for boundary points by gradually increasing the search direction towards $\bm{\hat{\eta}}_1$ in the plane spanned by $(\bm{\hat{v}}_1, \hat{\bm\eta}_1)$.  Figure~\ref{fig:boundary} displays the boundary in the 2-D plane using a reddish curved line in which the shaded region indicates the adversarial region. We notice that the non-targeted attack has a low curvature boundary, as opposed to the targeted attack, which has a high curvature boundary and a narrow adversarial region. Because of this difference in the boundary trajectory, CGBA outperforms CGBA-H for non-targeted attacks while CGBA-H outperforms CGBA for targeted attacks.

\begin{table}[t] \small
    \centering
    \begin{adjustbox}{width=0.47\textwidth}
    \fontsize{7}{7.8}\selectfont
    \begin{tabular}{c|c|c @{\hspace{0.85\tabcolsep}} c @{\hspace{0.85\tabcolsep}} c @{\hspace{0.85\tabcolsep}} c||  c| @{\hspace{0.85\tabcolsep}} c @{\hspace{0.85\tabcolsep}} c @{\hspace{0.85\tabcolsep}} c @{\hspace{0.85\tabcolsep}} c| }
    \toprule
    & \multicolumn{5}{c||}{  Non-targeted for different queries} & \multicolumn{5}{c|}{Targeted for different queries}\\ \hline
    $f$ &  & 1000 & 2500 & 5000 & 10000 &  & 1000 & 2500 & 5000 & 10000 \\
    \hline \hline
         4/3 & \multirow{4}{*}{\rotatebox[origin=c]{90}{SurFree}} &  8.93 & 4.79 & 2.57 & 1.62  & \multirow{4}{*}{\rotatebox[origin=c]{90}{AHA}} & 58.58 & 41.73 &  28.08 & 14.56\\
         2 &   & \textbf{8.44} & \textbf{4.42} & \textbf{2.65} &  \textbf{1.58} &  & 57.20 & 39.44 & 26.41 & 13.63\\
         4 &  & 9.88 & 5.26 & 3.02 & 1.71  &  & 55.85 & \textbf{37.36} & \textbf{23.14} & \textbf{11.44}\\
         8 &  & 9.29 & 4.86 & 2.85 & 1.73  &  & \textbf{53.59} & 37.61 & 24.18 & 14.48\\
         12 &  & 10.42 & 4.72 & 2.89 & 1.71 &  & 54.82 & 39.71 & 26.97 & 15.87\\\hline
         4/3 & \multirow{5}{*}{\rotatebox[origin=c]{90}{CGBA}}  & 10.49 & 4.09 & 2.07 & 1.19 &   \multirow{5}{*}{\rotatebox[origin=c]{90}{CGBA-H}} &  57.85 & 39.82 & 25.43 & 11.17 \\
         2 &   & 7.02 & 3.06 & 1.59 &  0.96 &  & 57.41 & 40.36 & 20.92 & 10.07 \\
         4 &   & 5.29 & 2.42 & 1.49 & \textbf{ 0.88} &  & \textbf{55.32} & 37.01 & \textbf{21.36} &  \textbf{9.57}\\
         8 &  & \textbf{4.72} & \textbf{2.35} & \textbf{1.43} & 1.05  &   & 55.62 & \textbf{36.77} & 21.79 & 10.08 \\
         12 &  & 4.81 & 2.39 & 1.65 & 1.41  &  & 55.54 & 39.63 & 26.61 & 15.94\\
         \bottomrule
    \end{tabular} 
    \end{adjustbox}  \vspace{-2mm} 
    \caption{Impact of dimension reduction on the performance.} \vspace{-3mm}
    \label{tab:dim_reductoin}
\end{table}

\paragraph{Impact of dimension reduction.} 
In Table~\ref{tab:dim_reductoin}, we compare the performance of CGBA and SOTA SurFree for the non-targeted attack, and the performance of CGBA-H and SOTA AHA for the targeted attack for different dimension reduction factor $f$. We randomly picked 100 images to perform the comparison of both attacks. For the non-targeted attack, SurFree offers the best performance for $f=2$. However, with the further increase of $f$, it does not show any performance improvement in dimension-reduced frequency subspace. 
In contrast, CGBA offers the best performance for $f=8$ with smaller query budgets and for $f=4$ with larger ones. In all cases, CGBA significantly outperforms SurFree in dimension-reduced frequency subspace. 
For the targeted attack, it is observed that the performance of AHA and CGBA-H is comparable for a limited query budget, but notably improved performance is obtained for CGBA-H with a sufficient query budget. 

\begin{figure} \vspace{-4mm}
    \centering
    \subfloat[]{\includegraphics[width=0.235\textwidth]{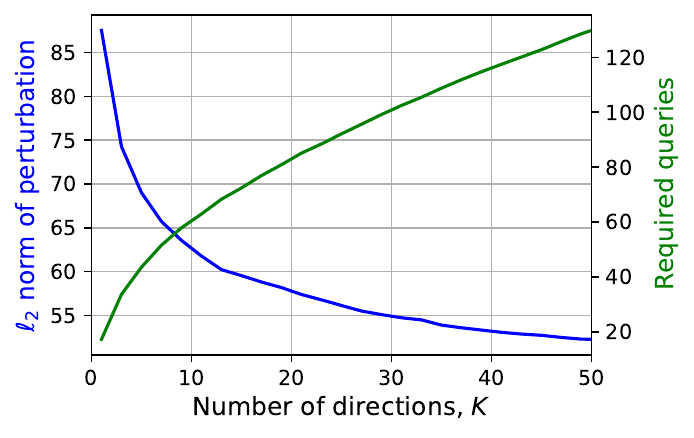}\label{init_tradeoff}}
    \subfloat[]{\includegraphics[width=0.23\textwidth]{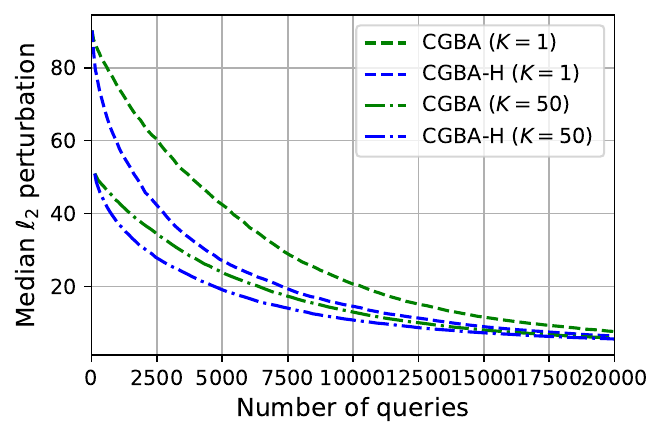}\label{radnInit_bestInit}} \vspace{-2mm}
    \caption{(a) Impact of the number of random direction $K$ to obtain $\bm x_{b_1}$; (b) Performance comparison between initialization with $K=1$ and $K=50$.}  \vspace{-4mm}
\end{figure}

\paragraph{Impact of initialization.} 
In the above experiments, the same random initialization was used for all methods for a fair comparison. In this part, we discuss the impact of initialization on the performance of proposed methods on targeted attacks as mentioned in \ref{init}. Figure~\ref{init_tradeoff} depicts the amount of perturbation and corresponding required queries to find $\bm x_{b_1}$ with different numbers of random directions by using $K$ samples of the target class. From this figure, a significant reduction in perturbation is observed with the increase of $K$. While with the random initialization, $K=1$, the obtained perturbation is around 85 by spending about 20 queries, a reduction in perturbation of more than 30 is obtained with $K=50$ by a small additional query cost of around 110. Figure~\ref{radnInit_bestInit} compares the performance of CGBA and CGBA-H with two different initialization: $K=1$ and $K=50$. Because a better initial boundary point is obtained by $K=50$ (with additional query cost properly counted), both CGBA and CGBA-H converge faster towards optimal perturbation than initialization with $K=1$. It's worth noting that the proposed initialization method can also be used to boost the baselines' performance. 

\section{Conclusion}
In this work, we have proposed two novel decision-based black-box attacks: CGBA and CGBA-H, which use a semicircular trajectory in a restricted 2D plane to ensure finding a new boundary point with reduced perturbation regardless of the boundary's curvature.
While CGBA outperforms the SOTA non-targeted attacks by effectively utilizing the low curvature of the decision boundary, CGBA-H is adapted to the high curvature of the decision boundary, resulting in better performance for targeted attacks.
Furthermore, we have introduced an initialization algorithm that can be used to find a better initial boundary point to further boost the performance for decision-based targeted attacks. We have conducted extensive experiments to verify the effectiveness of the proposed attacks.

\section*{Acknowledgements}
\vspace{-2mm} 
This work was supported in part by the US National Science Foundation under grants CNS-1824518 and ECCS-2203214. T. Wu is partly supported by ARO Grant W911NF1810295, NSF IIS-1909644, ARO Grant W911NF2210010, NSF IIS-1822477, NSF CMMI-2024688 and NSF IUSE-2013451. 
The views and conclusions contained herein are those of the authors and should not be interpreted as necessarily representing the official policies or endorsements, either expressed or implied, of the ARO, NSF, DHHS or the U.S. Government. The U.S. Government is authorized to reproduce and distribute reprints for Governmental purposes not withstanding any copyright annotation thereon. The authors are grateful for the constructive comments by anonymous reviewers and area chairs.

{\small
\bibliographystyle{ieee_fullname}
\bibliography{egbib}
}

\newpage
\clearpage
\section*{Supplementary Material}
\appendix

In this supplementary material, in Section~\ref{algorithms}, we illustrate the boundary search along a semicircular path (BSSP) and the Initialization algorithms. The experimental results on the ImageNet dataset are presented in Section~\ref{imagenet_results}, while those on the CIFAR10 dataset against two popular classifiers are given in Section~\ref{cifar_results}. Moreover, we compare our proposed BSSP over binary search along the direction of the estimated normal vector (BSNV) in Section~\ref{bsssp_bsnv} and verify its effectiveness through both theoretical analysis and experimental evaluation. Finally, we demonstrate and compare adversarial samples and their corresponding perturbations against different classifiers in Section~\ref{sec:visual_pert}.

\section{Algorithms} \label{algorithms}

The proposed CGBA is based on querying the boundary point along a semicircular path on a restricted 2D plane. The method to query a boundary point using BSSP is shown in Algorithm~\ref{alg:SemiCircularSearch}. This method is quite similar to the binary search. However, we conduct the boundary search on a semicircular path in a 2D plane rather than following a straight line 1D path. Starting from an adversarial and a non-adversarial point as two ends, we gradually decrease the range of the distance between adversarial and non-adversarial points on the semicircular trajectory to obtain the desired boundary point within a certain error limit. Line-3 and line-4 of the algorithm indicate the obtained unit directions $\hat{\bm \zeta}_c$ and $\hat{\bm \zeta}_{adv}$ towards a non-adversarial point $\bm x_c$ and an adversarial point $\bm x_{b_t}$ on the semicircular trajectory from source $\bm x_s$, respectively. Then, we obtain a perturbed point $\bm x_r = \bm x_s + \bm d(\hat{\bm \zeta}_r)$ on the semicircular trajectory in the resultant direction $\hat{\bm \zeta}_r$, obtained from the aforementioned directions as shown in line-6, where $d(\hat{\bm \zeta}_r)$ is the added perturbation to follow the semicircular trajectory as given in Eq.~\ref{eq:pert_semicircle}. From line-8 to line-11, the query for $\bm x_r$ is performed to know whether $\bm x_r$ is adversarial or not. If $\bm x_r$ is adversarial, $\hat{\bm \zeta}_{r}$ is replace by $\hat{\bm \zeta}_{adv}$ to reduce the search range, and vice-versa. The process of reducing the range is continued until we obtain the desired $\bm x_{b_{t+1}}$ within a certain accuracy.

\begin{algorithm}[t]
\small	
	\DontPrintSemicolon
	 \textbf{Inputs:} Source image $\bm x_s$, indicator function $\phi(.)$, non-adversarial point $\bm x_{c}$ ($\phi(\bm x_{c})=-1$) on the semicircle, adversarial sample at the intersection of the boundary and the semicircle $\bm x_{b_{t}}$, tolerance $\epsilon=0.0001$.
	 
	 \textbf{Output:} new boundary point $\bm x_{b_{t+1}}$.

	$\hat{\bm \zeta}_{c} = (\bm x_{c} - \bm x_s) / \|(\bm x_{c} - \bm x_s)\|_2$ \Comment*[r]{direction of a non-adversarial point $\bm x_c$ on the semicircle from $\bm x_s$}
	$\hat{\bm \zeta}_{adv} = (\bm x_{b_t} - \bm x_s) / \|(\bm x_{b_t} - \bm x_s)\|_2$ \Comment*[r]{direction of an adversarial point $\bm x_{b_t}$ on the semicircle from $\bm x_s$}
	\While{True}{
	    $\hat{\bm \zeta}_{r} = (\hat{\bm \zeta}_{c} + \hat{\bm \zeta}_{adv})/\|(\hat{\bm \zeta}_{c} + \hat{\bm \zeta}_{adv})\|_2$  
     
	    $\bm x_{r} = \bm x_s + \bm d(\hat{\bm \zeta}_{r})$ \Comment*[r]{to obtain $\bm x_r$ on the semicircle towards $\hat{\bm \zeta}_{r}$}

        \eIf{$\phi(\bm x_{r}) = 1$}{
            $\hat{\bm \zeta}_{adv} = \hat{\bm \zeta}_{r}$
        }{
        $\hat{\bm \zeta}_{c} = \hat{\bm \zeta}_{r}$
        }
 
		\If{$\|\bm d(\hat{\bm \zeta}_{adv}) - \bm d(\hat{\bm \zeta}_{c})\|_2 \leq \epsilon$}{
		$\bm x_{b_{t+1}} = \bm x_s + \bm d(\hat{\bm \zeta}_{adv})$\;
		break
        }
	} 
	\caption{BSSP}
	\label{alg:SemiCircularSearch} 
\end{algorithm}


\begin{algorithm}[h]
\small	
	\DontPrintSemicolon
	 \textbf{Inputs:} Source image $\bm x_s$, a set of directions towards the adversarial region $\{\bm \Theta_{k}\}_{k=1}^K$, indicator function $\phi(.)$ of target classier output.
	 
	 \textbf{Output:} Initial boundary point $\bm x_{b_1}$.

    $r \gets \min \{r>0: \phi(\bm x_s + r*\frac{\bm \Theta_{1}}{\| \bm \Theta_{1} \|_2}) =1\}$ \Comment*[r]{to find the minimum perturbation towards $\frac{\bm \Theta_{1}}{\| \bm \Theta_{1} \|_2}$ to make $\bm x_s$ adversarial} 
	 
	$\bm x_{b} = \bm x_s + r*\frac{\bm \Theta_{1}}{\|\bm \Theta_{1} \|_2}$\\
	$d_{best} = \|\bm x_{b} - \bm x_s\|_2$\\
	\For{$i=2:K$}{
	$\bm x_p = \bm x_s + d_{best}*\frac{\bm \Theta_{i}}{\|\bm \Theta_{i} \|_2} $\\
    	\If{$\phi(\bm x_p) =1 $}{ 
    	$\bm x_{b_1} \gets BinarySearch(\bm x_s, \bm x_p, \phi$)\\
    	$d_{new} = \|\bm x_{b_1} - \bm x_s\|_2$\\
        	\If{$d_{new} < d_{best}$}{
        	$d_{best}=d_{new}$
        	}
	}
	}
	\caption{Initialization}
	\label{alg:Initialization}
\end{algorithm}

\begin{figure}[h] \vspace{0mm}
    \centering
    \subfloat[VGG16]{\includegraphics[width=0.48\linewidth]{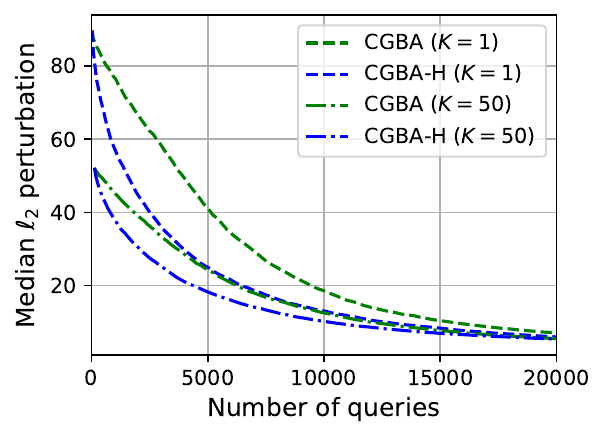}} \hfill
    \subfloat[ResNet101]{\includegraphics[width=0.48\linewidth]{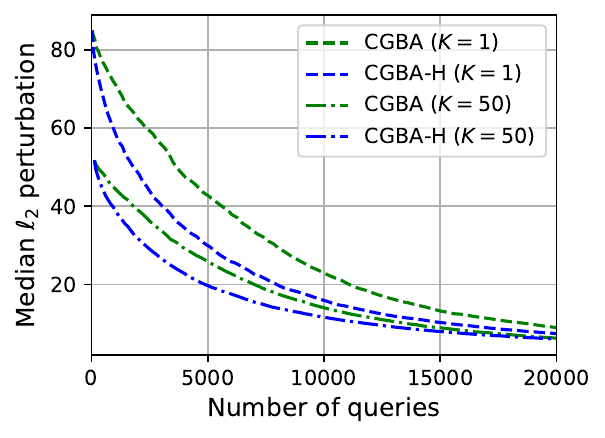}} \hfill
    \subfloat[ViT]{\includegraphics[width=0.48\linewidth]{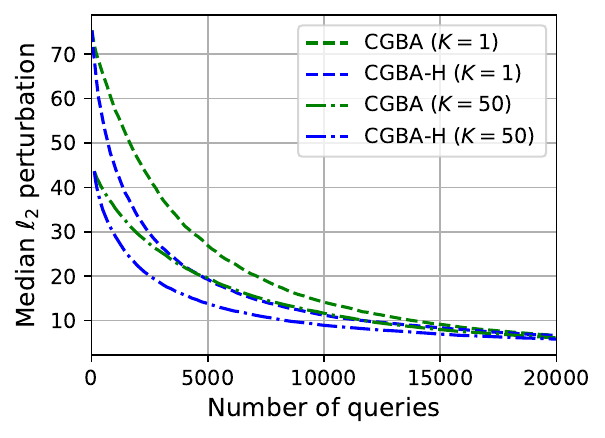}} \hfill
    \caption{{Performance comparison between a random initialization and the proposed initialization.}}
    \label{fig:all_ranInint_best}
\end{figure}

Algorithm \ref{alg:Initialization} shows the process of finding a better initial boundary point from a set of random directions to the adversarial region. If $\bm x_{k}$ denotes any point in the adversarial region, then the direction of $\bm x_{k}$ from a source $\bm x_s$ can be estimated as $\bm \Theta_k = (\bm x_{k} - \bm x_s)/\|\bm x_{k} - \bm x_s\|_2$. For the targeted attack, we randomly choose a set of $K$ samples $\{\bm x_k\}_{k=1}^K$ of the target class and obtain a set of $K$ directions $\{\bm \Theta_k\}_{k=1}^K$ to the adversarial region. By using this set of random directions, we can get a better initial boundary point $\bm x_{b_1}$ at the cost of additional queries. 
We provide the explanation of the Algorithm~\ref{alg:Initialization} as follows.

While the line-3 of Algorithm~\ref{alg:Initialization} finds the minimum $\ell_2$-norm of perturbation required to make $\bm x_s$ adversarial towards $\frac{\bm \Theta_{1}}{\| \bm \Theta_{1} \|_2}$, line-4 finds the adversarial image in that direction. The line-6 to line-12 is used to conduct an exhaustive search to find the direction that offers the best initial boundary point among the $K$ directions. In finding the best initial boundary point, for a direction $\frac{\bm \Theta_{i}}{\| \bm \Theta_{i} \|_2}$, line-7 adds the current perturbation $d_{best}$ towards $\frac{\bm \Theta_{i}}{\| \bm \Theta_{i} \|_2}$ to obtain a perturbed $\bm x_p$. Then, line-8 checks whether $\bm x_p$ is adversarial or not. If $\bm x_p$ is adversarial, only then we conduct a binary search to obtain a new boundary point, as shown in line-9, that further improves the obtained perturbation $d_{best}$. This process is continued to get the best boundary point among the given $K$ directions. 
Figure~\ref{fig:all_ranInint_best} compares the random initialization ($K=1$) and initialization with Algorithm~\ref{alg:Initialization} for $K=50$ against VGG16, ResNet101 and ViT classifiers. It is observed that the proposed initialization algorithm finds a much better boundary point against the aforementioned classifiers and thus notably improves the performance for targeted attacks.

\begin{table*}[]
\scriptsize
    \centering
\begin{tabular}{c|c|c c c c c c c|c c c c c c c|}
    \toprule
      \multicolumn{2}{r|}{Attack}  & \multicolumn{7}{c|}{Non-targeted} & \multicolumn{7}{c|}{Targeted}\\
      \hline
    \multicolumn{2}{r|}{Queries}  & 1000 & 2500 & 5000 & 7500 & 10000 & 15000 & 20000 & 1000 & 2500 & 5000 & 7500 & 10000 &15000 & 20000\\
    \midrule \midrule

    \multirow{7}{*}{\rotatebox[origin=c]{90}{ResNet101}} & HSJA~\cite{chen2019hopskipjumpattack} & 16.12 & 7.59 & 4.17 & 3.26 & 2.66 & 2.07 & 1.77 & 68.80 & 55.78 & 38.48 & 28.24 & 20.68 & 12.99 & 9.45\\ 
    & GeoDA~\cite{geoda} & 8.85 & 4.99 & 3.83 & 3.04 & 2.79 & 2.38 & 2.22 & - & - & - & - & - & - & - \\
    & TA~\cite{ma2021tangent} & 16.75 & 7.95 & 4.34 & 3.13 & 2.64 & 2.01 & 1.80 & 62.59 & 46.97 & 33.06 & 23.62 & 18.31 & 12.11 & 8.96\\
    & TriA~\cite{wang2022Triangle}  & 7.83 & 6.35 & 5.89 & 5.56 & 5.23 & 5.03 & 4.87 & - & - & - & - & - & - & -\\
    & SurFree~\cite{maho2021surfree}  & 10.47 & 5.62 & 3.12 & 2.16 & 1.79 & 1.35 & 1.11 & - & - & - & - & - & - & -\\
    & AHA~\cite{li2021aha} & - & - & - & - & - & - & - & 56.47 & 39.67 & 23.78 & 16.02 & 12.61 & 9.52 & 8.94\\ 
    & CGBA  & 7.89 & 3.38 & 1.84 & \textbf{1.25} & \textbf{1.02} & \textbf{0.77} & \textbf{0.66} & 73.17 & 60.38 & 39.85 & 25.47 & 17.48 & 9.26 & 6.13\\
    & CGBA-H & \textbf{7.19} & \textbf{3.32} & \textbf{1.79} & 1.26 & 1.03 & 0.78 & 0.67 & \textbf{55.69} & \textbf{37.28} &\textbf{ 21.59} & \textbf{14.32} & \textbf{10.77} & \textbf{6.55} & \textbf{4.52}\\

    \midrule
    \multirow{7}{*}{\rotatebox[origin=c]{90}{ViT}} & HSJA\cite{chen2019hopskipjumpattack} & 26.41 & 10.33 & 5.87 & 4.61 & 3.86 & 3.16 & 2.74 & 61.84 & 42.54 & 27.07 & 19.39 & 15.05 & 10.71 & 8.34\\
    & GeoDA~\cite{geoda} & 15.39 & 8.05 & 5.84 & 4.73 & 4.25 & 3.66 & 3.38 & - & - & - & - & - & - & -\\
    & TA~\cite{ma2021tangent} & 28.14 & 10.85 & 6.30 & 4.69 & 4.00 & 3.25 & 2.82 & 49.82 & 34.77 & 23.04 & 17.48 & 14.01 & 10.60 & 8.55\\
    & TriA~\cite{wang2022Triangle}  & 8.86 & 7.06 & 6.24 & 6.04 & 6.04 & 5.85 & 5.65 & - & - & - & - & - & - & -\\
    & SurFree~\cite{maho2021surfree}  & 14.96 & 6.90 & 4.11 & 3.10 & 2.53 & 1.95 & 1.61 & - & - & - & - & - & - & -\\
    & AHA~\cite{li2021aha}  & - & - & - & - & - & - & - & 43.06 & 28.29 & 17.54 & 12.59 & 9.58 & 6.77 & 5.74\\
    & CGBA  & 10.62 & 3.53 & \textbf{1.83} & \textbf{1.32} & \textbf{1.10} & \textbf{0.89} & \textbf{0.78} & 60.03 & 42.35 & 24.46 & 14.90 & 10.06 & 6.36  & 5.48\\
    & CGBA-H & \textbf{8.35} & \textbf{3.41} & {1.86} & 1.38 & 1.15 & 0.92 & 0.83 & \textbf{42.52} & \textbf{27.26} & \textbf{16.15} & \textbf{11.42} & \textbf{8.84} & \textbf{6.14} & \textbf{4.83}\\
    
     \bottomrule 
\end{tabular}   \vspace{-1mm}
    \caption{Median $\ell_2$-norm of perturbation for different query budgets against ResNet101 and ViT on ImageNet dataset.}
    \label{tab:norm_budget_resnet101_vit} \vspace{-3mm}
\end{table*}

\begin{figure}
    \centering
    \subfloat[Non-targeted attack]{\includegraphics[width=0.235\textwidth]{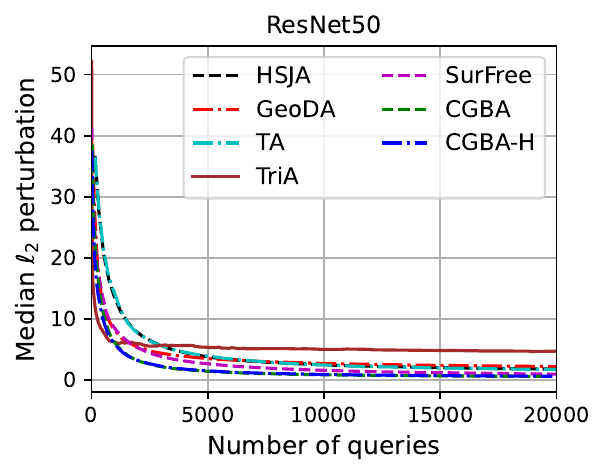}} \hfill
    \subfloat[Targeted attack]{\includegraphics[width=0.235\textwidth]{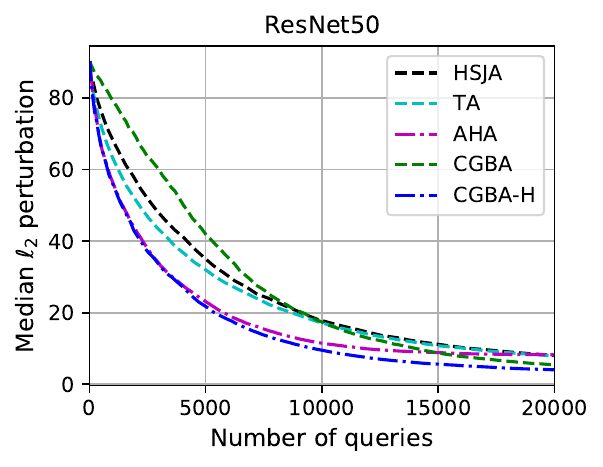}} \hfill

    \subfloat[Non-targeted attack]{\includegraphics[width=0.235\textwidth]{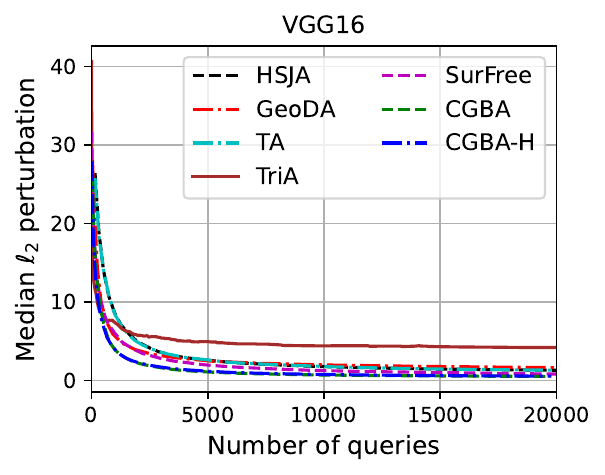}} \hfill
    \subfloat[Targeted attack]{\includegraphics[width=0.235\textwidth]{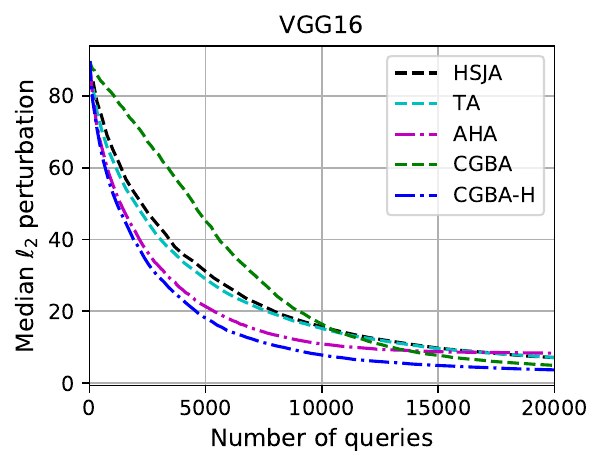}} \hfill

    \subfloat[Non-targeted attack]{\includegraphics[width=0.235\textwidth]{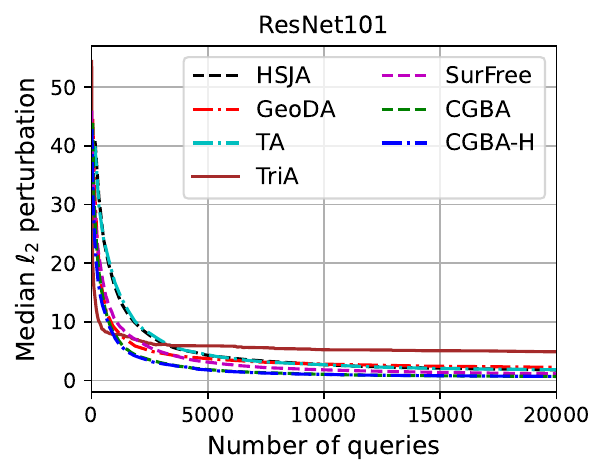}} \hfill
    \subfloat[Targeted attack]{\includegraphics[width=0.235\textwidth]{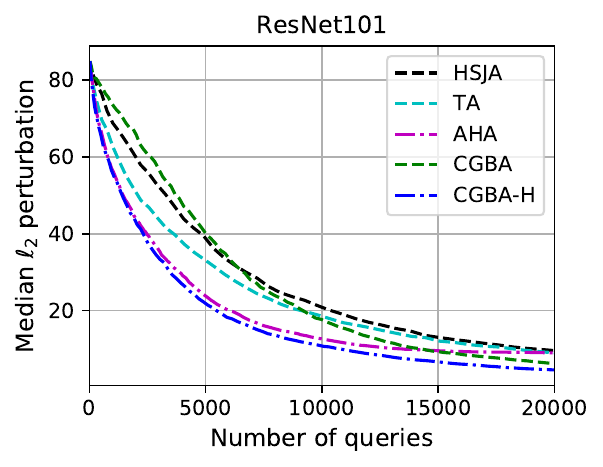}} \hfill

    \subfloat[Non-targeted attack]{\includegraphics[width=0.235\textwidth]{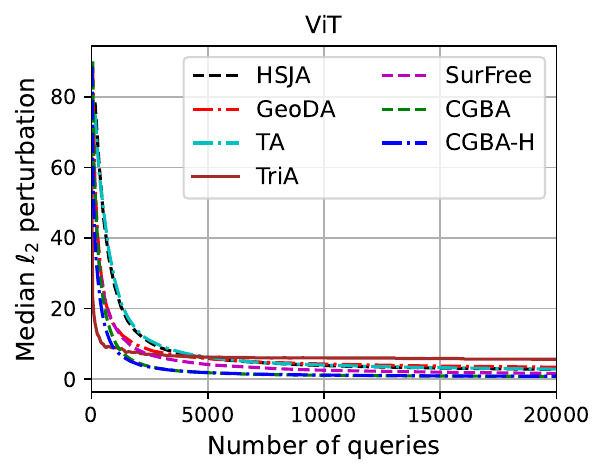}} \hfill
    \subfloat[Targeted attack]{\includegraphics[width=0.235\textwidth]{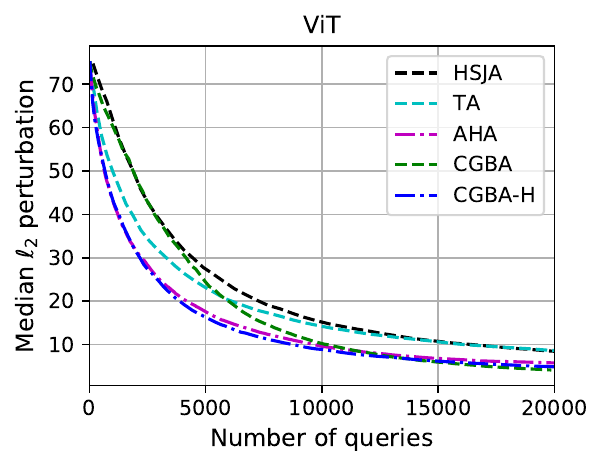}}
    \caption{Variation of $\ell_2$-norm of perturbation with the number of queries against ResNet50, VGG16, ResNet101 and ViT on ImageNet.}
    \label{fig:norm_queries}
\end{figure}

\begin{figure*}
    \subfloat[Non-targeted]{\includegraphics[width=0.23\textwidth]{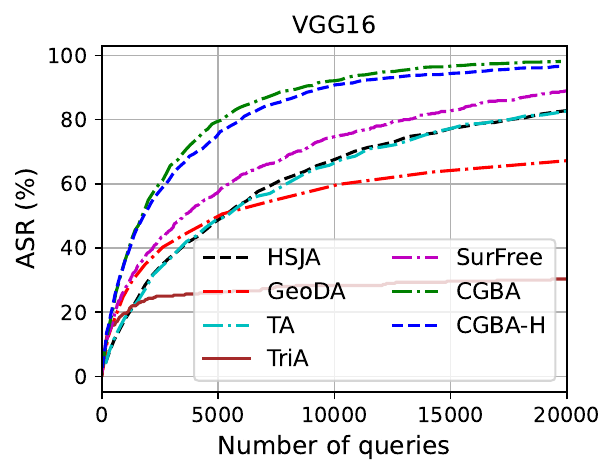}} \hfill
    \subfloat[Targeted]{\includegraphics[width=0.23\textwidth]{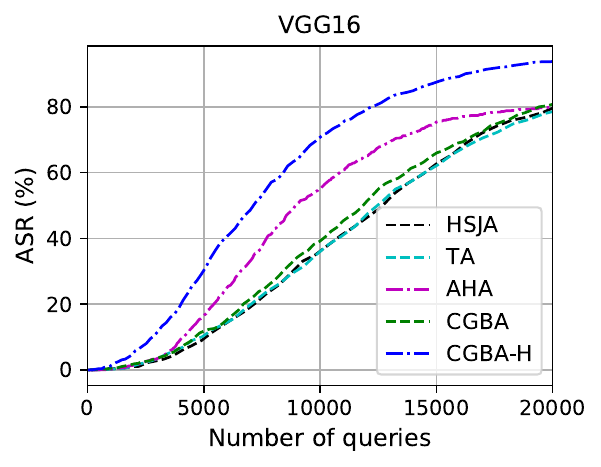}} \hfill
    \subfloat[Non-targeted]{\includegraphics[width=0.23\textwidth]{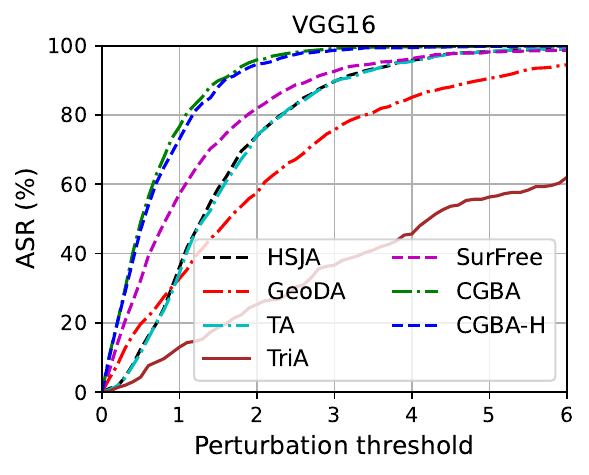}} \hfill
    \subfloat[Targeted]{\includegraphics[width=0.23\textwidth]{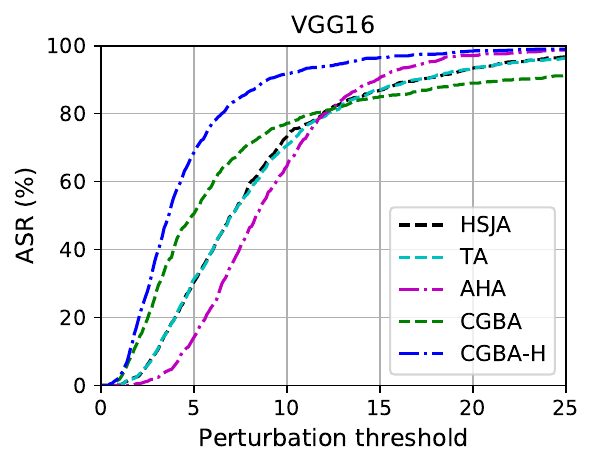}} \hfill

    \subfloat[Non-targeted]{\includegraphics[width=0.23\textwidth]{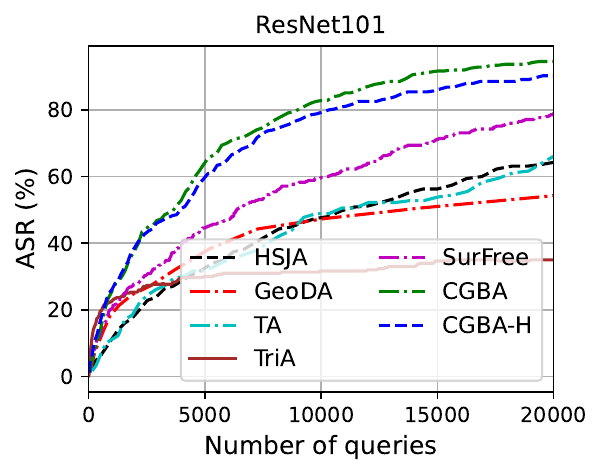}} \hfill
    \subfloat[Targeted]{\includegraphics[width=0.23\textwidth]{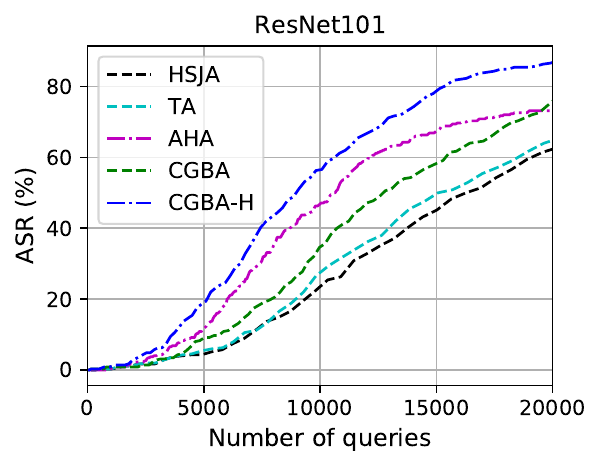}} \hfill
    \subfloat[Non-targeted]{\includegraphics[width=0.23\textwidth]{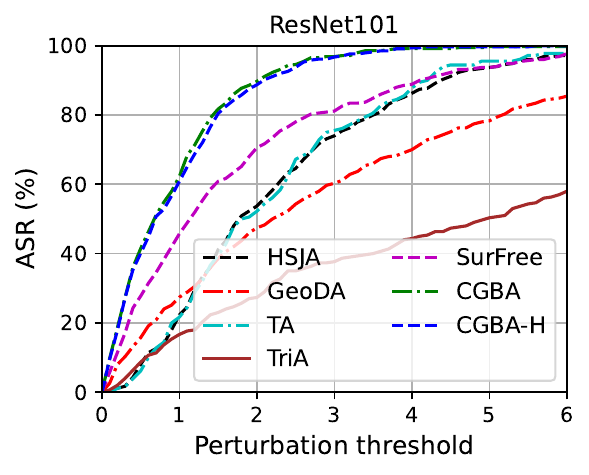}} \hfill
    \subfloat[Targeted]{\includegraphics[width=0.23\textwidth]{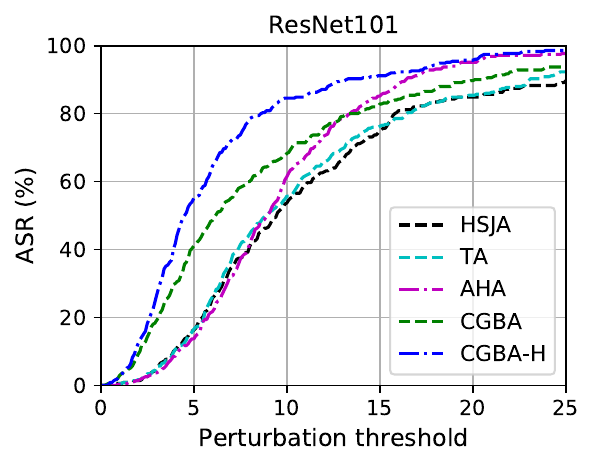}} \hfill

    \subfloat[Non-targeted]{\includegraphics[width=0.23\textwidth]{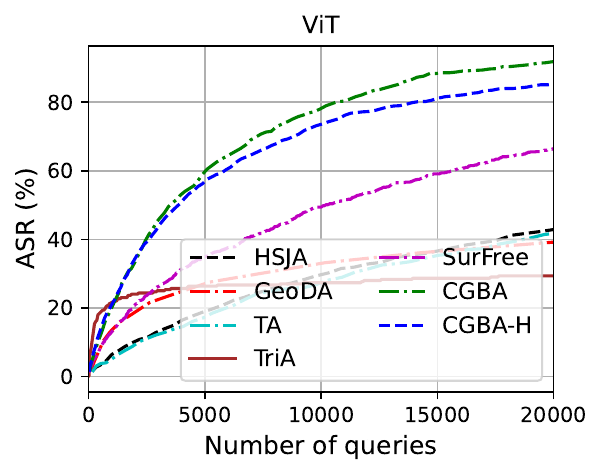}} \hfill
    \subfloat[Targeted]{\includegraphics[width=0.23\textwidth]{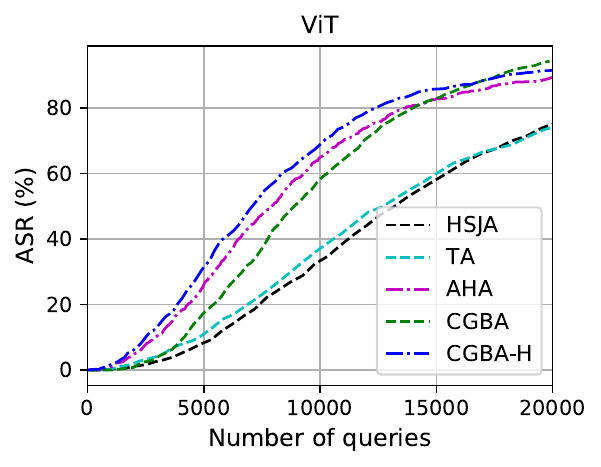}} \hfill
    \subfloat[Non-targeted]{\includegraphics[width=0.23\textwidth]{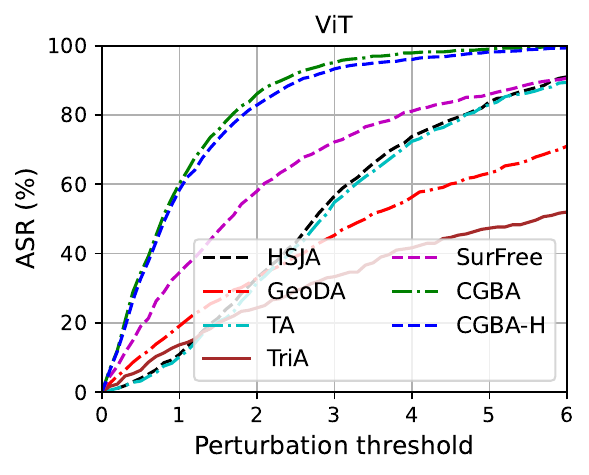}} \hfill
    \subfloat[Targeted]{\includegraphics[width=0.23\textwidth]{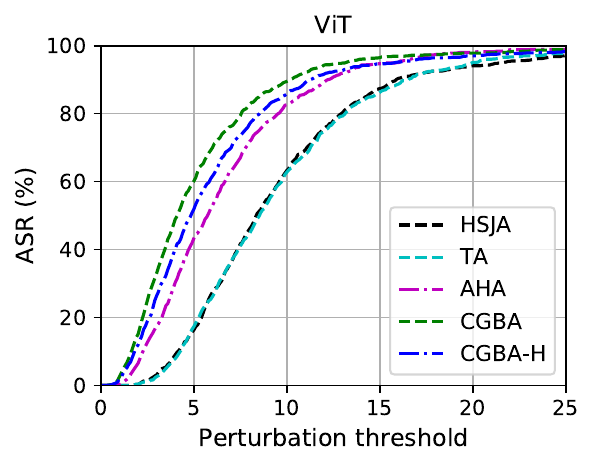}}

    \caption{Experimental results of ASR against VGG16, ResNet101 and ViT on ImageNet using different methods.}
    \label{fig:asr_all} 
\end{figure*}

\begin{table*}[t]
\footnotesize
    \centering
\begin{tabular}{l|c| c @{\hspace{1.2\tabcolsep}} c @{\hspace{1.2\tabcolsep}} c @{\hspace{1.2\tabcolsep}} c @{\hspace{1.2\tabcolsep}} c @{\hspace{1.2\tabcolsep}} c @{\hspace{1.2\tabcolsep}} c| c @{\hspace{1.2\tabcolsep}} c @{\hspace{1.2\tabcolsep}} c @{\hspace{1.2\tabcolsep}} c @{\hspace{1.2\tabcolsep}} c @{\hspace{1.2\tabcolsep}} c @{\hspace{1.2\tabcolsep}} c|}
    \toprule
      & & \multicolumn{7}{c|}{Non-targeted} & \multicolumn{7}{c|}{Targeted }\\
      \hline
    & Queries & 1000 & 2500 & 5000 & 7500 & 10000 & 15000 & 20000 &  1000 & 2500 & 5000 & 7500 & 10000 &15000 & 20000\\
    \midrule \midrule
    \multirow{7}{*}{\rotatebox[origin=c]{90}{PreActResNet18}} 
    & HSJA~\cite{chen2019hopskipjumpattack}  & 0.531 & 0.279 & 0.203 & 0.174 & 0.161 & 0.146 & 0.140 & 1.771 & 0.682 & 0.421 & 0.337 & 0.304 & 0.268 & 0.252\\
    & GeoDA~\cite{geoda}  & 2.41 & 1.87 & 1.526 & 1.349 & 1.296 & 1.176 & 1.116 & - & - & - & - & - & -  & -\\
    & TA~\cite{ma2021tangent} & 0.502 & 0.274 & 0.199 & 0.178 & 0.166 & 0.150 & 0.144 & 1.657 & 0.670 & 0.408 & 0.337 & 0.301 & 0.272  & 0.256\\
    & TriA~\cite{wang2022Triangle}  & 0.621 & 0.504 & 0.469 & 0.436 & 0.433 & 0.414 & 0.406 & - & - & - & - & - & -  & -\\
    & SurFree~\cite{maho2021surfree}   & 0.428 & 0.270 & 0.204 & 0.177 & 0.163 & 0.147 & 0.140 & - & - & - & - & - & -  & -\\
    & AHA~\cite{li2021aha} & - & - & - & - & - & - & - & 4.096 & 1.968 & 1.126 & 1.053 &  1.053 & 1.053 & 1.053 \\
    & CGBA  & \textbf{0.409} & \textbf{0.237} & \textbf{0.184} & \textbf{0.163} & \textbf{0.152} & \textbf{0.140} & \textbf{0.135} & 2.012 & 0.645 & 0.391 & \textbf{0.321} & \textbf{0.291} & \textbf{0.262}  & \textbf{0.247}\\
    & CGBA-H   & 0.435 & 0.267 & 1.95 & 0.172 & 0.159 & 0.145 & 0.140 & \textbf{1.257} & \textbf{0.577} & \textbf{0.385} & 0.329 & 0.296 &  0.266 & 0.251\\
 
     \midrule
     
    \multirow{7}{*}{\rotatebox[origin=c]{90}{WRN40}} 
    & HSJA~\cite{chen2019hopskipjumpattack} & 0.739 & 0.336 & 0.206 & 0.166 & 0.148 & 0.129 & 0.123 & 4.028 & 1.315 & 0.596 & 0.410 & 0.336 & 0.271  & 0.244\\
    & GeoDA~\cite{geoda} & 3.241 & 2.243 & 1.677 & 1.491 & 1.387 & 1.264 & 1.162 & - & - & - & - & - & - & -\\
    & TA~\cite{ma2021tangent} & 0.714 & 0.334 & 0.209 & 0.169 & 0.149 & 0.132 & 0.125 & 3.736 & 1.229 & 0.571 & 0.396 & 0.330 &  0.272 & 0.250\\
    & TriA~\cite{wang2022Triangle}  & 0.949 & 0.697 & 0.625 & 0.574 & 0.541 & 0.528 & 0.501 & - & - & - & - & - & -  & -\\
    & SurFree~\cite{maho2021surfree}  & \textbf{0.493} & 0.262 & 0.187 & 0.156 & 0.142 & 0.127 & 0.119 & - & - & - & - & - & -  & \\
    & AHA~\cite{li2021aha} & - & - & - & - & - & - & - & 5.372 & 2.709 & 1.359 & 1.081 & 1.044 & 1.041 & 1.041\\
    & CGBA  & 0.498 & \textbf{0.245} & \textbf{0.167} & \textbf{0.142} & \textbf{0.131} & \textbf{0.120} & \textbf{0.115} & 6.221 & 1.774 & 0.578 & 0.383 & 0.312 & \textbf{0.256}  & \textbf{0.231}\\
    & CGBA-H   & 0.537 & 0.259 & 0.172 & 0.148 & 0.135 & 0.122 & {0.116} & \textbf{2.690} & \textbf{0.878} & \textbf{0.465} & \textbf{0.351} & \textbf{0.302} &  0.257 & 0.238\\
     \bottomrule
\end{tabular}
    \caption{Median $\ell_2$-norm of perturbation for different query budgets of our proposed attacks and baselines on CIFAR10 dataset.}
    \label{tab:norm_query_cifar}
\end{table*}

\begin{figure*} [t]
\centering
    \subfloat[Non-targeted attack]{\includegraphics[width=0.235\textwidth]{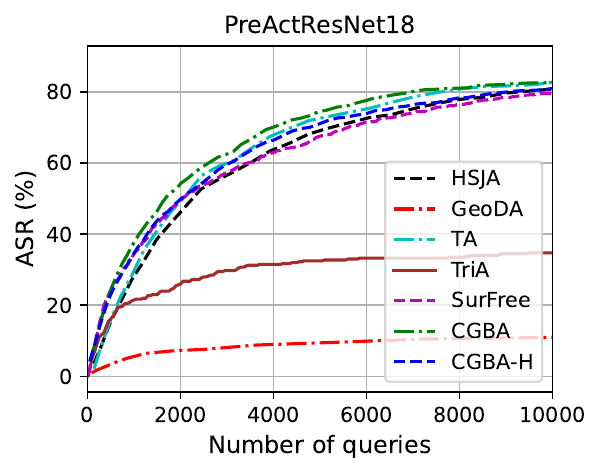}} \hfill
    \subfloat[Targeted attack]{\includegraphics[width=0.235\textwidth]{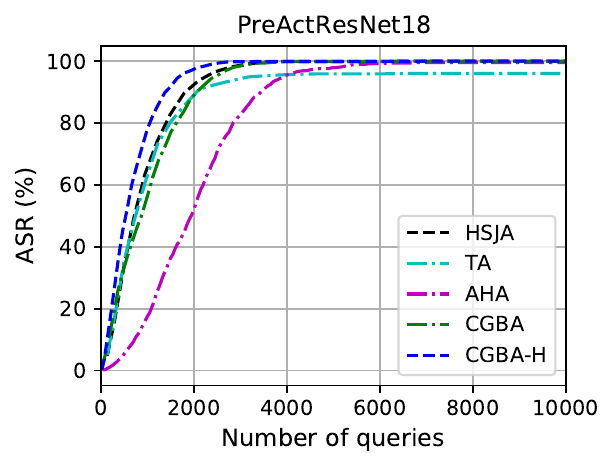}} \hfill
    \subfloat[Non-targeted attack]{\includegraphics[width=0.235\textwidth]{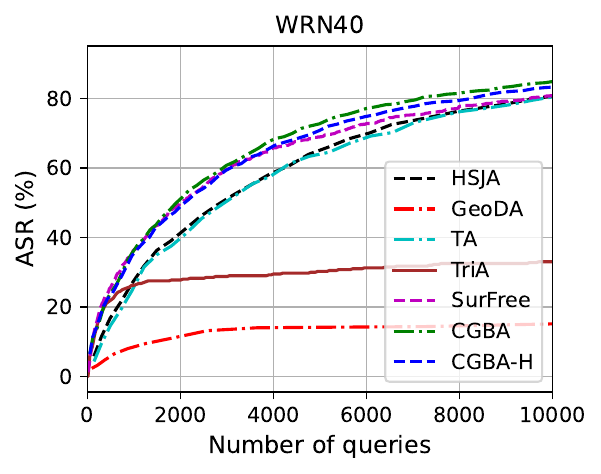}} \hfill
    \subfloat[Targeted attack]{\includegraphics[width=0.235\textwidth]{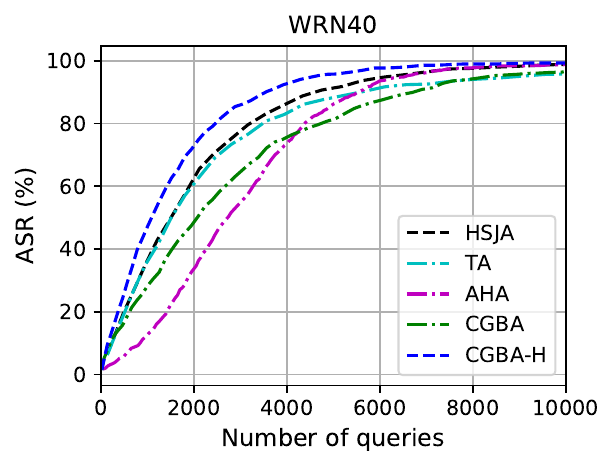}}

    \subfloat[Non-targeted attack]{\includegraphics[width=0.235\textwidth]{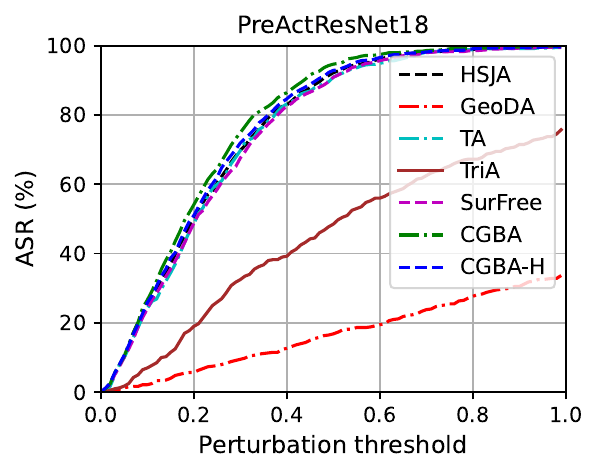}} \hfill
    \subfloat[Targeted attack]{\includegraphics[width=0.235\textwidth]{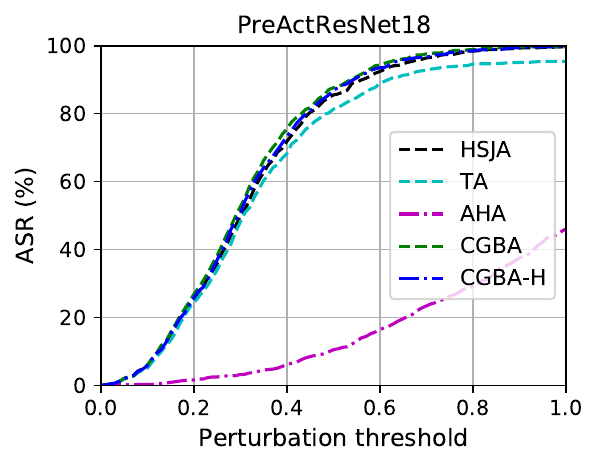}} \hfill
    \subfloat[Non-targeted attack]{\includegraphics[width=0.235\textwidth]{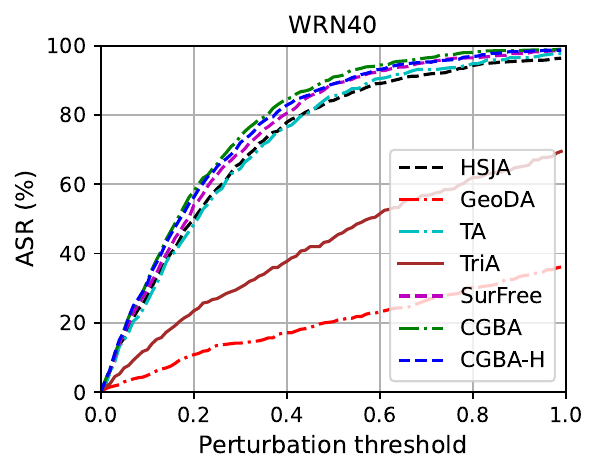}} \hfill
    \subfloat[Targeted attack]{\includegraphics[width=0.235\textwidth]{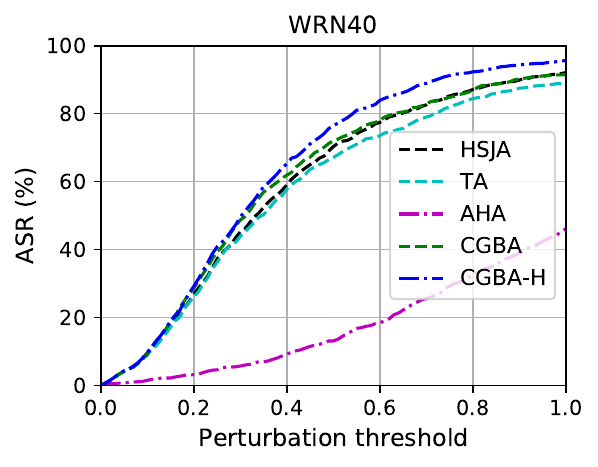}}

    \caption{Experimental results of ASR against PreActResNet18 and WRN40 on CIFAR10 using different methods.}
    \label{fig:asr_query_cifar}
\end{figure*}

\begin{table*} [h]
\small
\centering
    \begin{tabular}{c|c|c  @{\hspace{1.15\tabcolsep}} c @{\hspace{1.15\tabcolsep}} c @{\hspace{1.15\tabcolsep}} c @{\hspace{1.15\tabcolsep}} c @{\hspace{1.15\tabcolsep}} c @{\hspace{1.15\tabcolsep}} c @{\hspace{1.35\tabcolsep}} c|}
    \toprule
        &Methods & HSJA~\cite{chen2019hopskipjumpattack} & GeoDA~\cite{geoda} & TA~\cite{ma2021tangent} & TriA~\cite{wang2022Triangle} & SurFree~\cite{maho2021surfree} & AHA~\cite{li2021aha} & CGBA & CGBA-H  \\ 
        \midrule
        \multirow{2}{*}{{PreActResNet18}} & Non-targeted & 3281.2 & 16007 &  3293.7 & 5216.4 & 2867.5  & -  &  \textbf{2693.8} &  2866.0 \\
        & Targeted & 9056.2  & -  &  8882.3 & - & -  &  19478.3 &  9742.0 &  \textbf{7734.3   }\\
        \hline
        \multirow{2}{*}{{WRN40}} & Non-targeted & 3704.9 & 19119 & 3876.5 & 7025.9 & 2881.4 & - & \textbf{2860.3} & 2948.9 \\
        & Targeted & 14322.7 & - & 13806.6 & - & - & 23543.7 &  18443.9 &  \textbf{11376.2} \\
        
        \bottomrule
    \end{tabular} \vspace{-2mm}
    \caption{AUC comparison against PreActResNet18 and WRN40 for a query budget of 10000 on CIFAR10.}
    \label{tab:auc_cifar}
\end{table*}

\section{Results on ImageNet} \label{imagenet_results}
Figure~\ref{fig:norm_queries} depicts the variation of median $\ell_2$-norms of perturbation with queries for both non-targeted and targeted attacks against ResNet50, VGG16, ResNet101 and ViT. Additionally, Table~\ref{tab:norm_budget_resnet101_vit} presents the obtained perturbation for different query budgets against ResNet101 and ViT. Based on these findings, it is evident that for non-targeted attacks, both CGBA and CGBA-H outperform the baseline methods significantly. In contrast, for targeted attacks, while CGBA-H outperforms all the baselines, the obtained perturbation using CGBA is expectedly higher when the query budget is not sufficiently high due to the high curvature of the boundary. However, with the increase in the number of iterations and corresponding queries, the boundary point gets closer to the source image, resulting in a flatter decision boundary with respect to the viewpoint of the source image. Consequently, with a sufficient query budget, the proposed CGBA also outperforms the baselines. 

Figure~\ref{fig:asr_all} shows the attack success rate (ASR) comparison of the proposed methods with the baselines for the different query budgets and threshold values. The first two columns of the figure depict the obtained ASR for different query budgets with a threshold value of 2.5 for the non-targeted attack and 12 for the targeted attack, while the last two columns show the obtained ASR for different threshold values for a query budget of 20,000. From these experimental results, we observe that CGBA and CGBA-H offer significantly better performance over the baselines for these three popular classifiers, as we have observed for ResNet50.

\section{Results on CIFAR10} \label{cifar_results}
Our proposed attacks are not restricted to high-dimensional datasets with a large number of classification labels, such as ImageNet. This section further examines their effectiveness in generating adversarial samples for a low-dimensional dataset, CIFAR10, which contains ten classification labels.
To perform the experiments, rather than using the dimension-reduced subspace, we consider full-dimensional image space (full-space) to attack against PreActResNet18~\cite{he2016identity} and WRN40~\cite{zagoruyko2016WRN} using the baselines and the proposed methods. We randomly choose 1000 images for the non-targeted attack and 1000 pairs of images for the targeted attack that are correctly classified by the target classifier. 
The corresponding results are shown in Table.~\ref{tab:norm_query_cifar}. From this Table, CGBA outperforms all the baselines, as expected, for the non-targeted attack against PreActResNet18. On the other hand, for the non-targeted attack against WRN40, while SurFree performs better with a very low query budget, CGBA outperforms SurFree with a sufficient query budget. However, for the targeted attack, while CGBA-H performs better with smaller query budgets, CGBA shows slightly better performance than CGBA-H with an increase in the query budget. This observation could be explained as follows. First of all, since the number of classes in the CIFAR10 dataset is much smaller, the adversarial region for the targeted attack on CIFAR10 is much wider in the 2D search plane as compared to the adversarial region of the ImageNet dataset with 1000 classes. Secondly, with the increase in the number of queries, the obtained boundary point is getting closer to the source image. Thus, the curvature of the boundary becomes flatter from the viewpoint of the source. Therefore, with a sufficiently large query budget, which in turn requires a large number of iterations, CGBA performs better on a low-dimensional dataset like CIFAR10. This is consistent with our previous observations that CGBA performs better on lower curvature boundaries while CGBA-H can further adapt to high curvature.

Moreover, the obtained ASR against the two classifiers is shown in Figure~\ref{fig:asr_query_cifar}. Figures~\ref{fig:asr_query_cifar}(a-d) demonstrate the variation of ASR with queries for a perturbation threshold of 0.3 for the non-targeted attack and 2.5 for the targeted attack. On the other hand, Figures~\ref{fig:asr_query_cifar}(e-h) demonstrate the obtained ASR with different threshold values. To depict Figures~\ref{fig:asr_query_cifar}(e-h), while we consider a query budget of 5000 for the non-targeted attack, we consider a query budget of 10000 for the targeted attack. We choose two different query budgets due to the faster convergence of the non-targeted attack than the targeted attack. From these figures, it is observed that we get the expected improved performance using our proposed methods.   
Furthermore, the experimental results of AUC are shown in Table~\ref{tab:auc_cifar}. The obtained results demonstrate the consistency in the performance of the proposed methods on different datasets.

\section{BSSP versus BSNV} \label{bsssp_bsnv}
In this section, we compare binary search along the direction of the estimated normal vector (BSNV)  and boundary search along a semicircular path (BSSP) to find a new boundary point. First, we show the experimental evaluation of these two methods in finding the new boundary point with the queries spent to estimate the normal vector on the boundary point. Then we provide a theoretical analysis to further justify the improved efficiency of BSSP in comparison with BSNV.
\subsection{Impact of normal vector estimation}
\begin{figure}
    \centering
    \includegraphics[width=0.3\textwidth]{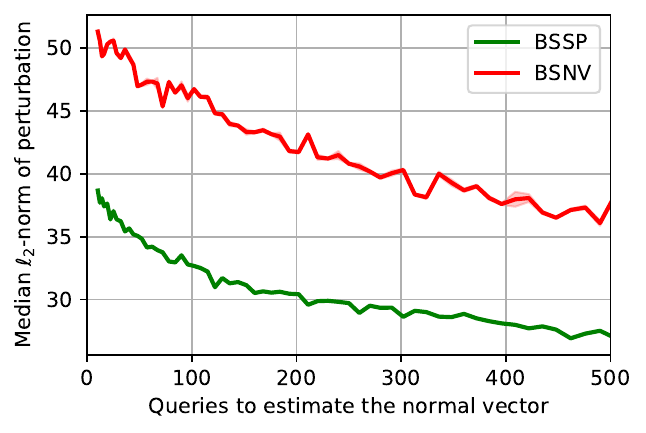}
    \caption{Impact of the normal vector estimation on the performance of BSSP and BSNV  in finding a new boundary point.}
    \label{fig:impact_of_normalEstimation}
\end{figure}
We demonstrate the impact of the normal vector estimation on the performance of BSSP and BSNV in finding a new boundary point in Figure \ref{fig:impact_of_normalEstimation}. To depict this figure, we consider the non-targeted attack against ResNet50 ~\cite{he2016deep} for 1000 test images from ImageNet~\cite{deng2009imagenet}. First, we estimate the normal vector at the same initial boundary point $\bm x_{b_1}$ considering different numbers of queries. Then, we use both BSNV and the proposed BSSP to find a new boundary point and compare the difference in the resultant perturbations. It is clearly seen that while the performance of both methods improves with the increase of queries spent for the normal vector estimation as expected, BSSP consistently outperforms BSNV by a large margin.

\subsection{Theoretical analysis}
\begin{figure}[t]
    \centering
    \includegraphics[trim = 2mm 2mm 2mm 2mm, clip,width=0.30\textwidth]{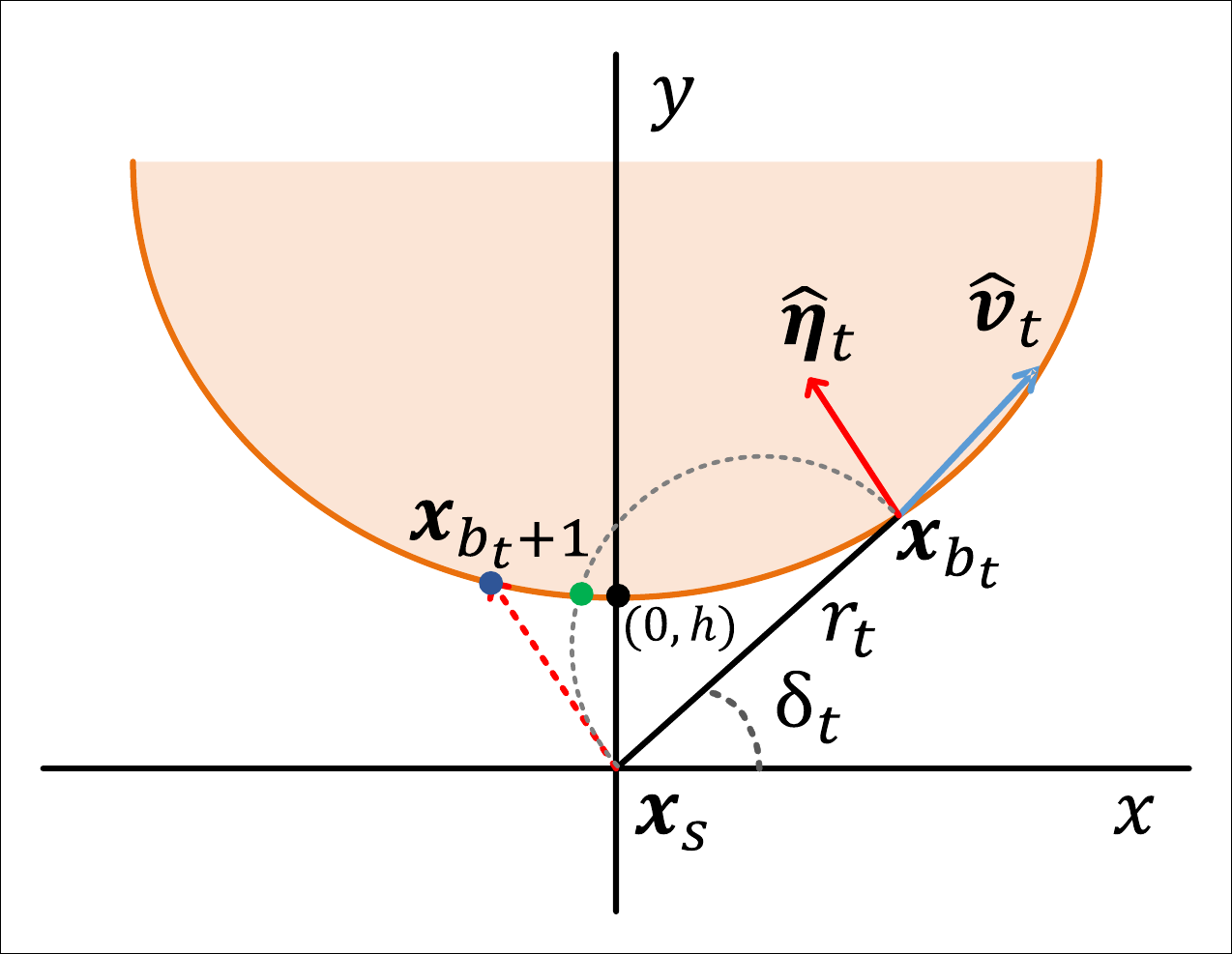}
    \caption{A parabolic boundary with vertex at $(0,h)$, where $\bm x_{b_t}$ represents the boundary point at $t$-th iteration, and the green and blue points on the boundary denote the obtained boundary point $\bm x_{b_{t+1}}$ using BSSP and BSNV, respectively.}
    \label{fig:parabolic_boundary}
\end{figure}

We theoretically verify the advantage of the BSSP algorithm compared to BSNV  in finding a new boundary point. As BSSP is conducted in a 2D plane, we consider a hypothetical parabolic boundary in the 2D plane to perform our analysis for tractability. 
Let the source image $\bm x_s$ be located at the origin of a $xy$-coordinate plane spanned by $(\hat{\bm v}_t, \hat{\bm \eta}_t)$ at iteration $t$, as shown in Figure~\ref{fig:parabolic_boundary}, where $\hat{\bm v}_t$ is the direction of the boundary point $\bm x_{b_t}$ from source $\bm x_s$ and $\hat{\bm \eta}_t$ is the estimated normal vector at $\bm x_{b_t}$. Assume the boundary separating the benign and adversarial regions of $\bm x_s$ in the 2D plane is represented as a parabolic function:
\begin{equation}
    y = \frac{x^2}{4p}+h,
    \label{eq:parabola}
\end{equation}
whose coordinate of the vertex is at $(0,h)$ and the length of the latus rectum is $4p$. Therefore, the optimal perturbation required to make $\bm x_s$ adversarial is $h$ at the given iteration $t$.

Let us assume the direction of the boundary point $\bm x_{b_t}$ w.r.t. the $x$-axis is $\delta_t$ and the amount of perturbation in that particular direction is $r_t = \|\bm x_{b_t} - \bm x_s\|_2$ at the $t$-th iteration. Thus, the projection of perturbation in the direction of $x$-axis and $y$-axis is given as $a_{x_t} = r_t \cos{\delta_t}$ and $a_{y_t} = r_t \sin{\delta_t}$, respectively.
Hence, by putting the values of $a_{x_t}$ and $a_{y_t}$ in Eq.~\ref{eq:parabola}, $r_t$ is related to $\delta_t$ as

\begin{equation}
    r_t = \frac{2p \sin{\delta_t}}{\cos^2{\delta_t}} \left[1 - \sqrt{1 - \frac{h}{p}\cot^2{\delta_t}}\right].
    \label{eq:rt}
\end{equation}

As can be inferred from Eq.~\ref{eq:rt}, it is possible to find a boundary point at the direction $\delta_t$ iff $p \ge h \cot^2{\delta_t}$. Now using Eq.~\ref{eq:rt}, we have, 

\begin{equation} \label{eq:a_x}
    a_{x_t} = r_t \cos{\delta_t} = 2p \tan{\delta_t} \left[1 - \sqrt{1 - \frac{h}{p}\cot^2{\delta_t}}\right]
\end{equation}
\begin{equation} \label{eq:a_y}
    a_{y_t} = r_t \sin{\delta_t} = 2p \tan^2{\delta_t} \left[1 - \sqrt{1 - \frac{h}{p}\cot^2{\delta_t}}\right].
\end{equation}

In this analysis, we consider two extreme scenarios for both BSNV and BSSP: linear boundary and curved boundary such that the line towards $\hat{\bm v}_t$ is tangent with the boundary.

\subsubsection{Finding boundary point using BSNV }
The tangent of any point on the parabola is given by
\begin{equation}
    \frac{dy}{dx} = \frac{x}{2p}.
\end{equation}
Therefore, the slope of the line in the normal direction $\hat{\bm \eta}_t$ on the boundary point $(a_{x_{t}}, a_{y_{t}})$ can be expressed as
\begin{equation} \label{eq:m}
    m = -\frac{2p}{a_x}.
\end{equation}
Hence, the next boundary point $\bm x_{b_{t+1}}$ toward $\hat{\bm \eta}_t$ is located on the line $y=mx$. Let $(a_{x_{t+1}}, a_{y_{t+1}})$ be the coordinate of the boundary point $\bm x_{b_{t+1}}$ on the $xy$-coordinate plane. Then, $(a_{x_{t+1}}, a_{y_{t+1}})$ can be obtained at the intersection point of $y=mx$ and $y = \frac{x^2}{4p}+h$. Therefore, we have

\begin{equation}
    (a_{x_{t+1}}, a_{y_{t+1}}) = \bigg(\frac{2h/m}{1+ \sqrt{1- \frac{h}{pm^2}}}, \frac{2h}{1+ \sqrt{1- \frac{h}{pm^2}}}\bigg),
    \label{eq:next_bound_coordinate}
\end{equation}
where $m = -\frac{2p}{a_x}$.


\paragraph{Case 1: Linear boundary $(p=\infty)$.} For this case, pushing the image $\bm x_s$ towards the normal direction $\hat{\bm \eta}_t$, the Eq.~\ref{eq:next_bound_coordinate} will result in
\begin{equation}
    (a_{x_{t+1}}, a_{y_{t+1}}) = (0, h).
\end{equation}
{Thus, we can conclude that the BSNV method finds the subsequent boundary point with optimal perturbation, $r_{t+1} = h$, if the boundary is linear and the normal vector estimate is accurate.  This explains the success of qFool and GeoDA when the decision boundary can be well approximated as a plane.}

\begin{figure}[h]
    \centering
    \includegraphics[trim={3mm 4mm 3mm 4mm},clip, width=0.50\textwidth]{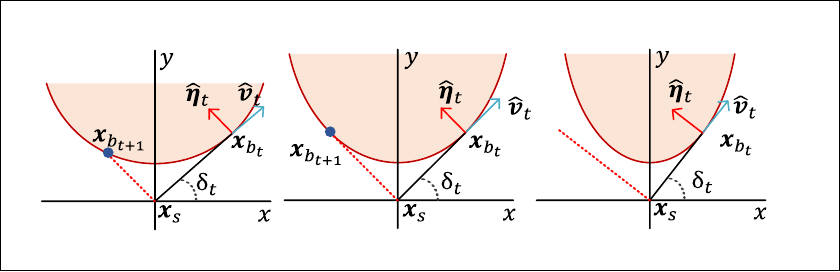}
    \caption{Obtaining a new boundary point $\bm x_{b_{t+1}}$ using BSNV when $\hat{\bm v}_t$ is tangent on the boundary at $\bm x_{b_t}$ for $\delta_t < 45^0$ (left), $\delta_t = 45^0$ (middle) and $\delta_t > 45^0$ (right). BSNV cannot find $\bm x_{b_{t+1}}$ if $\delta_t > 45^0$.}
    \label{fig:bsnv_diffCurve}
\end{figure}

\paragraph{Case 2: Curved boundary ($p = h \cot^2{\delta_t}$).} With this condition, the line starting from the source $\bm x_s$ with  an angle $\delta_t$ from the $x$-axis is the tangent with the boundary at $(a_{x_{t}}, a_{y_{t}})$. From Eq.~\ref{eq:a_x} and Eq.~\ref{eq:a_y}, we have $(a_x, a_y) = (2p \tan{\delta_t}, 2p \tan^2{\delta_t})$. Thus, from Eq.~\ref{eq:m}, we get the slope of line in the normal direction $\hat{\bm \eta}_t$ on the boundary point as
\begin{equation}
    m = - \frac{2p}{2p \tan{\delta_t}} = - \cot{\delta_t}.
    \label{eq:m_curved}
\end{equation}

Now, if we use the BSNV method to find $\bm x_{b_{t+1}}$, from Eq.~\ref{eq:next_bound_coordinate} we can find a valid boundary point $\bm x_{b_{t+1}}$, only if $pm^2 \ge h$. Therefore, using the condition $p = h \cot^2{\delta_t}$ and Eq.~\ref{eq:m_curved}, we get
\begin{equation}
    \delta_t \le 45^0.
\end{equation}

Thus, if the line in the direction of $\delta_t$ w.r.t. $x$-axis is the tangent on $\bm x_{b_t}$, the BSNV method will find the subsequent boundary point $\bm x_{b_{t+1}}$ only if $\delta_t \le 45^0$. Consider the extreme condition when $\delta_t = 45^0$, for which we have $m=-1$ and $p=h$. Therefore, from Eq.~\ref{eq:next_bound_coordinate}, the coordinate of the subsequent boundary point can be written as
\begin{equation}
    (a_{x_{t+1}}, a_{y_{t+1}}) = (-2h, 2h).
\end{equation}
The amount of perturbation of the obtained boundary point $\bm x_{b_{t+1}}$ is $r_{t+1} = 2\sqrt{2}h$, which is same as $r_t = 2\sqrt{2}p$, and the iterative querying process will not converge.
So, we can conclude that if $p = h \cot^2{\delta_t}$, finding the subsequent boundary point using the BSNV method converges iif $\delta_t < 45^0$ in this scenario, as shown in Figure~\ref{fig:bsnv_diffCurve}.

\subsubsection{Finding boundary point using BSSP}
In this subsection, we theoretically analyze the amount of perturbation required to make $\bm x_s$ adversarial by using the BSSP method to find the boundary point in the 2-D plane spanned by $(\hat{\bm v}_t, \hat{\bm \eta}_t)$. 
The boundary point $(a_{x_{t+1}}, a_{y_{t+1}})$ can be simply obtained by finding the intersection of the parabolic boundary given in Eq.~\ref{eq:parabola} and the circle specified by the following equation.
\begin{equation}
    (x - \frac{r_t}{2} \cos{\delta_t})^2 + (y - \frac{r_t}{2} \sin{\delta_t})^2 = \frac{r^2}{4} . \notag 
\end{equation}
Thus, we have
\begin{align}
    a_{x_{t+1}}^2 + (\frac{a_{x_{t+1}}^2}{4p} + h)^2 -  \frac{2h \cot{\delta_t}}{1 + \sqrt{1 - \frac{h}{p} \cot^2{\delta_t}}} a_{x_{t+1}} \notag \\ \quad -  \frac{2 h}{1 + \sqrt{1 - \frac{h}{p} \cot^2{\delta_t}}} (\frac{a_{x_{t+1}}^2}{4p}+h) = 0 . \label{eq:semi_intersection}
\end{align}

\paragraph{Case 1: Linear boundary $(p=\infty)$.} 
Under this condition, the coordinate of $\bm x_{b_{t+1}}$ can be calculated by solving the Eq.~\ref{eq:semi_intersection} as
\begin{equation}
    (a_{x_{t+1}}, a_{y_{t+1}}) = (0, h).
\end{equation}
{Hence, the BSSP also finds the optimal boundary point $\bm x_{b_{t+1}}$ with minimum perturbation $\|\bm x_{b_{t+1}} - \bm x_s\|_2 = h$  as it is obtained by using BSNV for a linear boundary.}

\begin{figure}[t]
    \centering
    \includegraphics[trim={3mm 4mm 3mm 5mm},clip, width=0.49\textwidth]{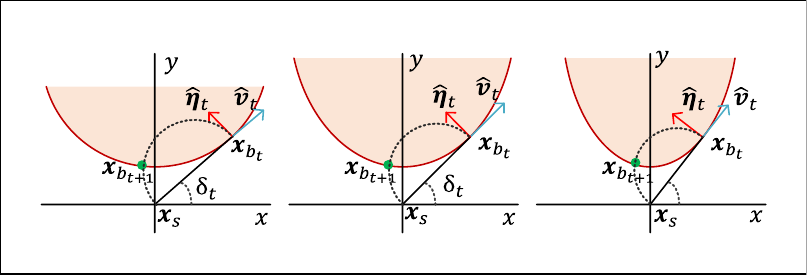}
    \caption{Obtaining a new boundary point $\bm x_{b_{t+1}}$ using BSSP when $\hat{\bm v}_t$ is tangent on the boundary at $\bm x_{b_t}$ for $\delta_t < 45^0$ (left), $\delta_t = 45^0$ (middle) and $\delta_t > 45^0$ (right). BSSP finds $\bm x_{b_{t+1}}$ irrespective of the boundary's curvature.}
    \label{fig:BSSP_diffCurve}
\end{figure}

\paragraph{Case 2: Curved boundary $(p=h \cot^2{\delta_t})$.} 
In this case, Eq.~\ref{eq:semi_intersection} can be written as
\begin{equation}
    \frac{a_{x_{t+1}}^4}{16h^2} + a_{x_{t+1}}^2- 2h \cot{\delta_t} a_{x_{t+1}}  + h^2 = 0.
\end{equation}
One solution of the above equation is the coordinate of the current boundary point $\bm x_{b_t}$. As the coefficients of the above equation are real, there must be another real solution irrespective of the value of $\delta_t$. Thus, for a given boundary point $\bm x_{b_t}$, the proposed BSSP method ensures finding the subsequent boundary point $\bm x_{b_{t+1}}$ irrespective of the value of $\delta_t$. As $p=h \cot^2{\delta_t}$ and the curvature is related to the latus rectum $4p$,  we can say, conversely, that the proposed BSSP is guaranteed to find the next boundary point no matter what the boundary curvature is, and it is depicted in Figure~\ref{fig:BSSP_diffCurve}. In contrast, as we have seen, BSNV cannot find a new boundary point when $\delta_t > 45^0$ under the condition of $p=h \cot^2{\delta_t}$.

As a concrete example, consider the extreme conditions that we consider for BSNV above, where the direction of $\bm x_{b_t}$ from $\bm x_s$ is a tangent at $\bm x_{b_t}$ and creates an angle $\delta_t = 45^0$. Therefore, Eq.~\ref{eq:semi_intersection} satisfying these conditions can be written as
\begin{align}
     \frac{a_{x_{t+1}}^4}{16h^2} + a_{x_{t+1}}^2- 2ha_{x_{t+1}}  + h^2 = 0.
    \label{eq:curved_boundary_bssp}
\end{align}
By solving Eq.~\ref{eq:curved_boundary_bssp}, we can obtain the coordinate of the subsequent boundary point $\bm x_{b_{t+1}}$ as
\begin{equation}
    (a_{x_{t+1}}, a_{y_{t+1}}) = (-0.4135h, 1.0427h).
\end{equation}
The perturbation of the new boundary point $\bm x_{b_{t+1}}$ is
\begin{equation}
    r_{t+1} = \sqrt{(-0.4135 h)^2 + (1.0427 h)^2} = 1.1217 h.  \notag
\end{equation}

Hence, under the conditions of  $p=h \cot^2{\delta_t} ~\text{and}~ \delta_t = 45^0$, the amount of reduction in perturbation using BSSP as compared to BSNV is obtained as
\begin{equation}
   \frac{2\sqrt{2}h - 1.1217h}{2\sqrt{2}h} = 60.3\%. \notag
\end{equation}


\begin{figure}
    \centering
    \subfloat[$h=10$ and $\delta_t=30^0$]{
    \includegraphics[width=0.20\textwidth]{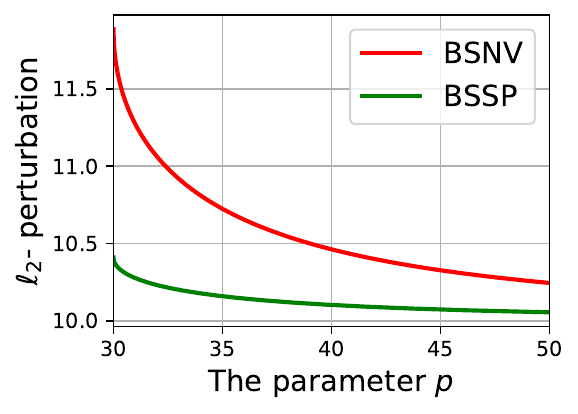}
    }
    \hfill
    \subfloat[$h=10$ and $\delta_t=45^0$]{
    \includegraphics[width=0.20\textwidth]{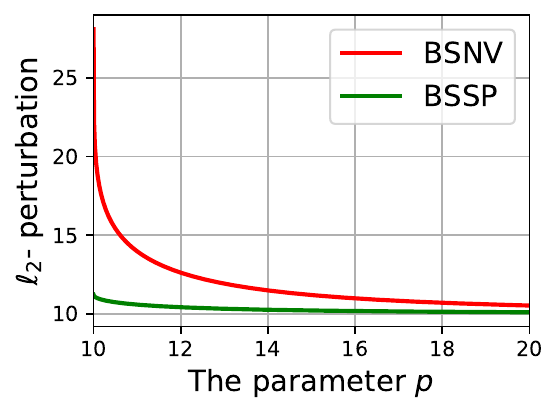}
    }
    \hfill
    \subfloat[$h=10$ and $\delta_t=60^0$]{
    \includegraphics[width=0.20\textwidth]{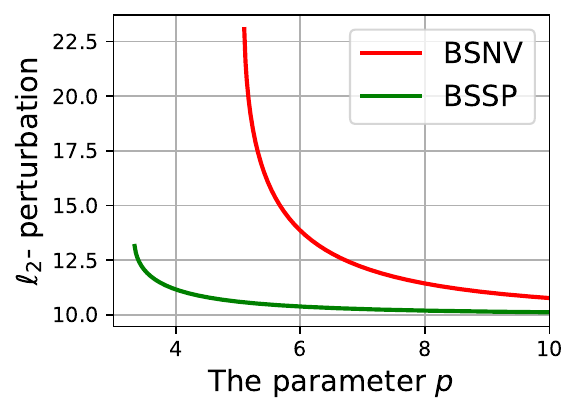}
    }
    \hfill
    \subfloat[$h=10$ and $\delta_t=80^0$]{
    \includegraphics[width=0.20\textwidth]{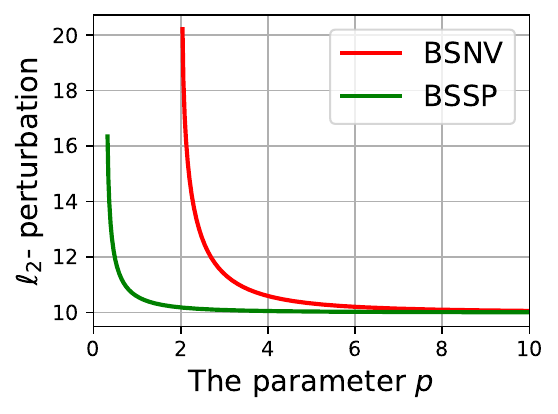}
    }
    \caption{Performance Comparision for different values of parameter $p$.}
    \label{fig:norm_p_parabolic_boundary}
\end{figure}

\subsubsection{Impact of different curvature}
We further investigate the impact of the boundary curvature on the performance of these two search algorithms. We have already seen that these two methods achieve the same optimal performance for the linear boundary ($p=\infty$). However, when $p=h \cos^2{\delta_t}$ and $\delta_t = 45^0$, BSSP can reduce the perturbation by about 60\%  as compared to BSNV. Figure~\ref{fig:norm_p_parabolic_boundary} shows the performance comparison of BSNV and BSSP for different values of $p$ (with smaller $p$ corresponding to higher curvature). We consider $h=10$ and $\delta_t=30^0, 45^0, 60^0 \text{ \& } 80^0$ to obtain the solutions of the aforementioned equations for different values of $p$ numerically. From this figure, we observe that BSSP uniformly outperforms BSNV, with dramatic improvement when the curvature is high. 
Moreover, as shown clearly in Figure~\ref{fig:norm_p_parabolic_boundary}(c,d), while BSNV fails when parameter $p$ falls below a certain threshold, BSSP remains functioning under such high curvature settings. 
\section{Visualizing perturbation} \label{sec:visual_pert}
\begin{figure*}[h]
    \centering
    \subfloat[ResNet50]{\includegraphics[width=0.83\textwidth]{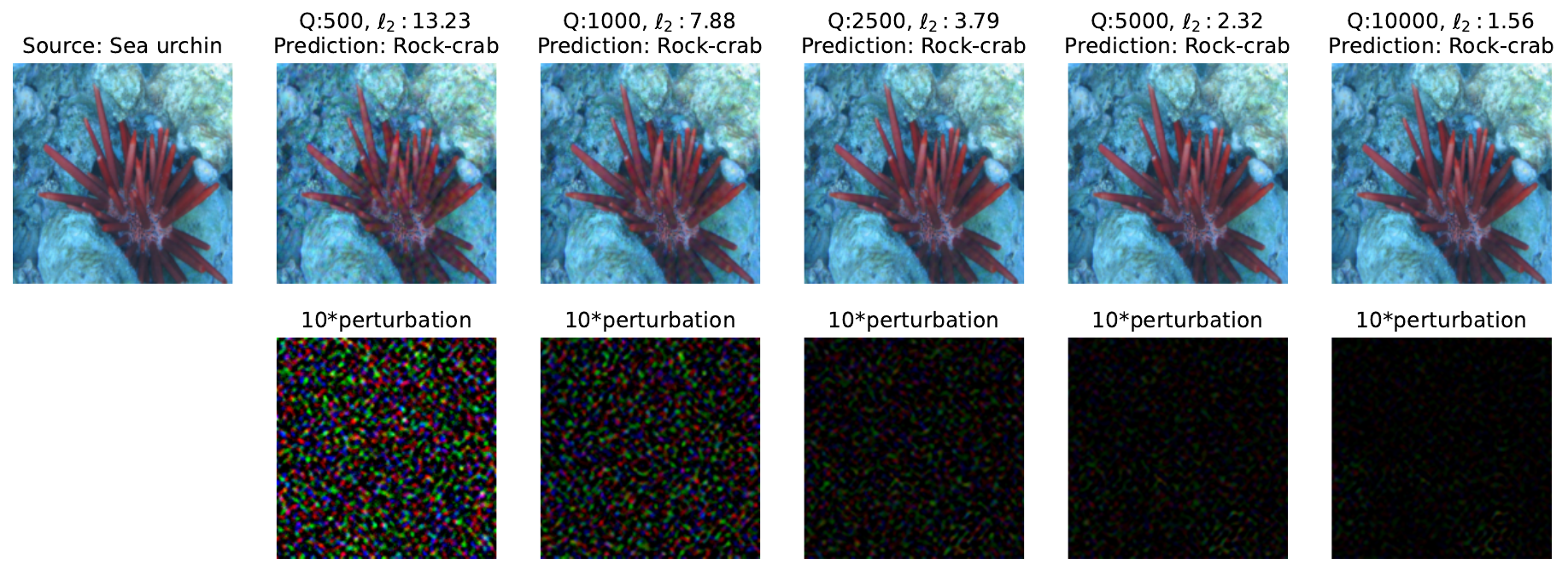}}
    \vspace{-2mm}
    \subfloat[VGG16]{\includegraphics[width=0.83\textwidth]{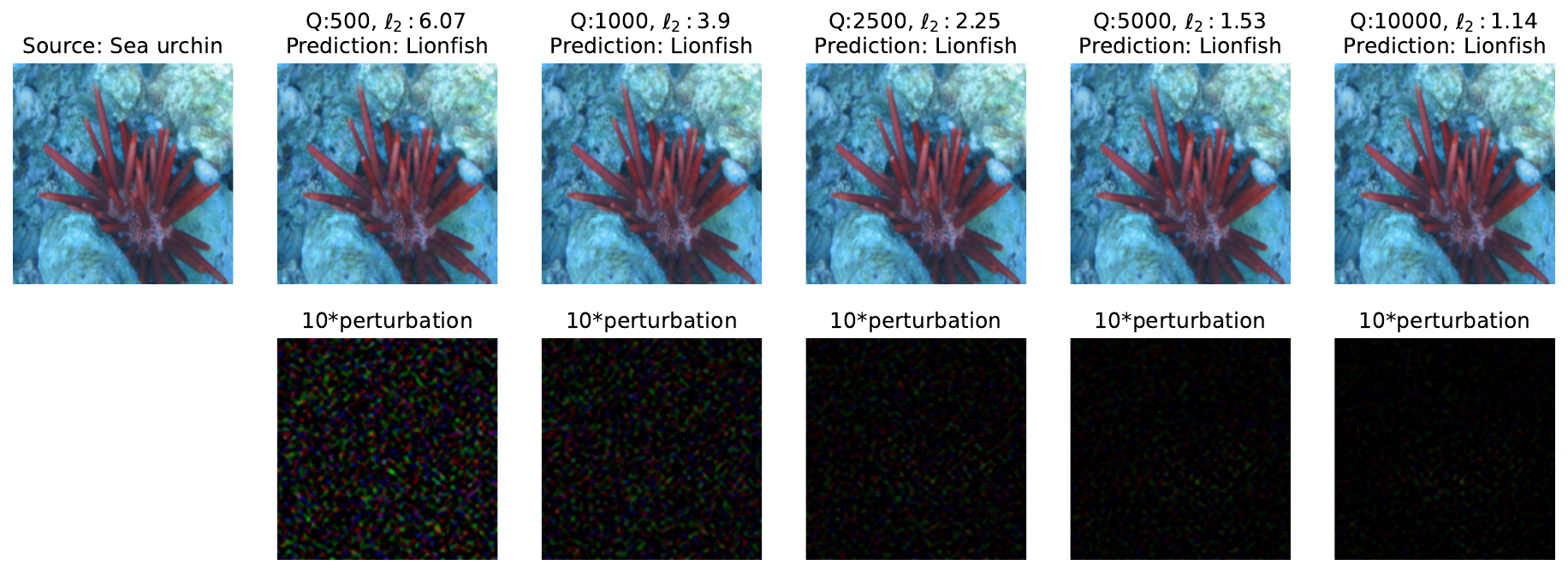}}
    \vspace{-2mm}
    \subfloat[ ViT]{\includegraphics[width=0.83\textwidth]{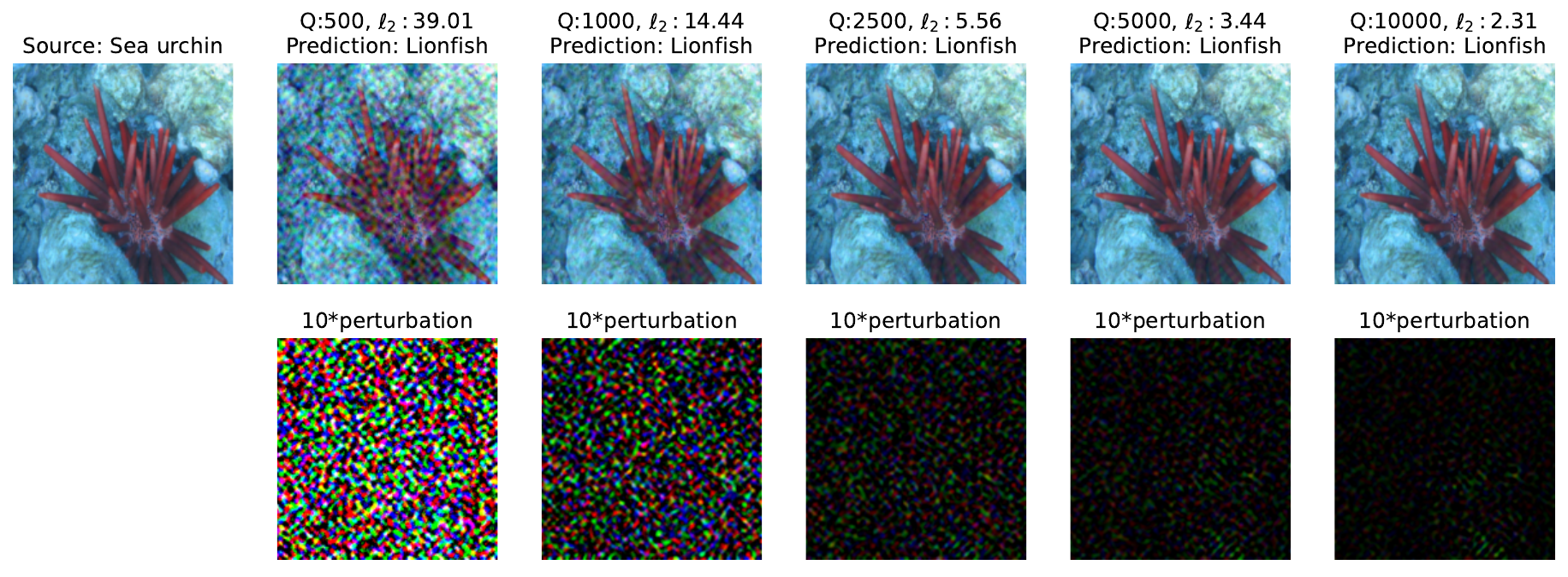}}
    \caption{Obtained adversarial images and corresponding amplified perturbations with different query budgets against different classifiers using non-targeted CGBA.}
    \label{fig:visul_adv_non_tar}
\end{figure*}

\begin{figure*}
    \centering
    \includegraphics[width=0.83\textwidth]{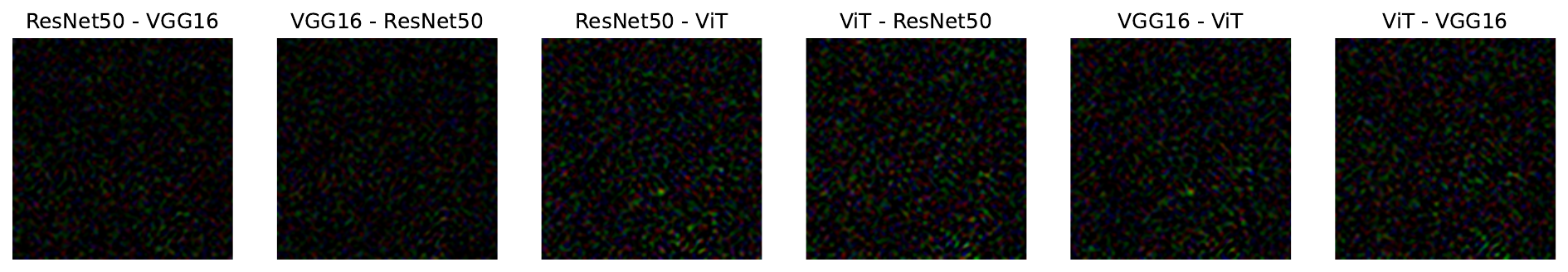}
    \caption{Difference in crafted amplified perturbation in Figure~\ref{fig:visul_adv_non_tar} between different classifiers for a query budget of 5000.}
    \label{fig:non_tar_pert_diff}
\end{figure*}

\begin{figure*}[h]
    \centering
    \subfloat[ResNet50]{\includegraphics[width=0.83\textwidth]{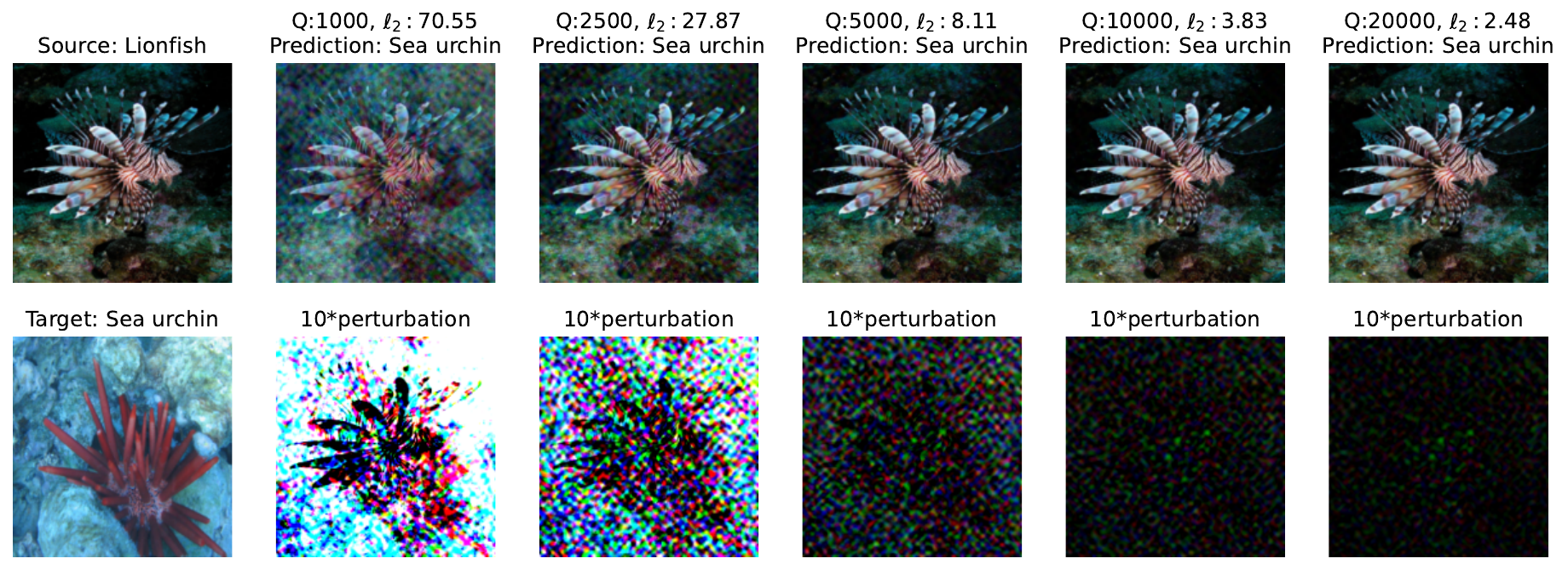}}
    \vspace{-2mm}
    \subfloat[VGG16]{\includegraphics[width=0.83\textwidth]{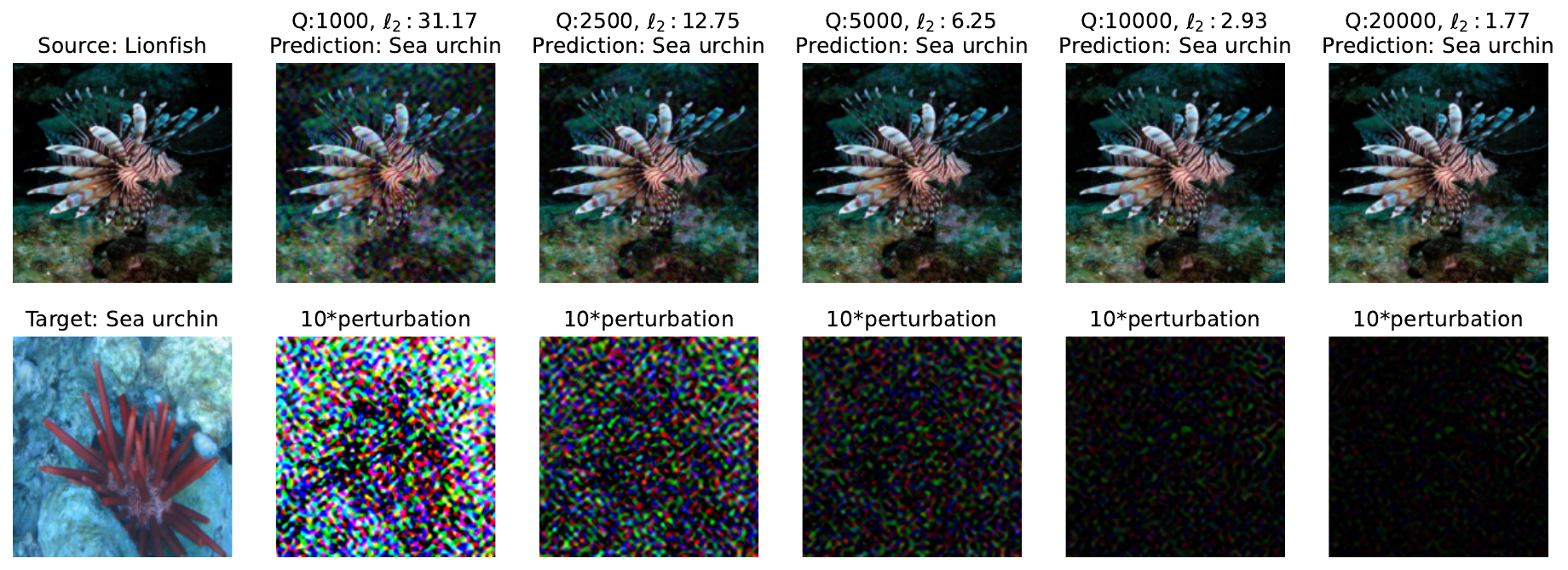}}
    \vspace{-2mm}
    \subfloat[ ViT]{\includegraphics[width=0.83\textwidth]{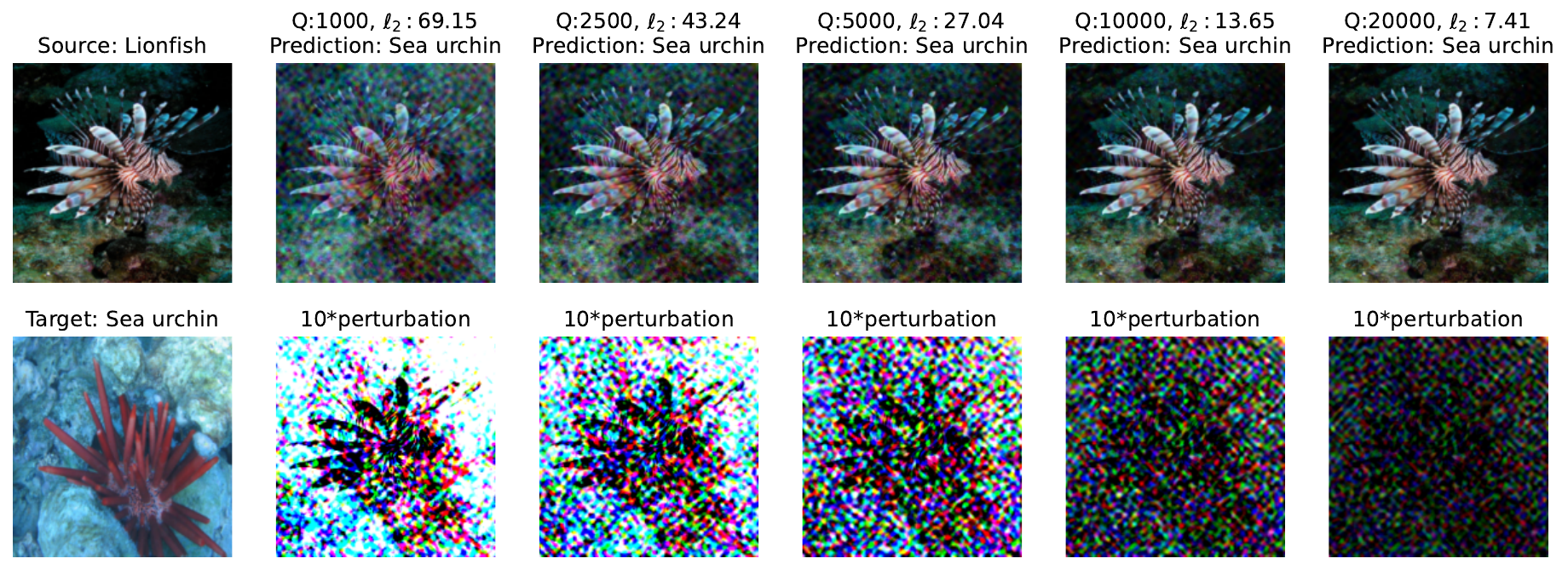}}
    \caption{Obtained adversarial images and corresponding amplified perturbations with different query budgets against different classifiers using targeted CGBA-H.}
    \label{fig:visul_adv_tar}
\end{figure*}

\begin{figure*}
    \centering
    \includegraphics[width=0.83\textwidth]{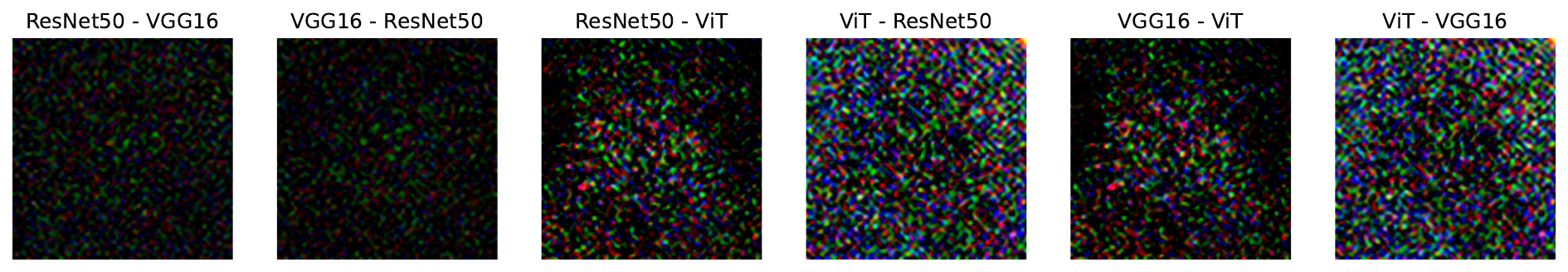}
    \caption{Difference in crafted amplified perturbation in Figure~\ref{fig:visul_adv_tar} between different classifiers for a query budget of 10000.}
    \label{fig:tar_pert_diff}
\end{figure*}

In this section, we visualize obtained adversarial images and corresponding perturbations for different query budgets for both non-targeted and targeted attacks against ResNet50, VGG16 and ViT. We use test images from the ILSVRC2012's validation set~\cite{deng2009imagenet} that are correctly classified by the target classifiers. In Figure~\ref{fig:visul_adv_non_tar}, the first row of each sub-figures demonstrates the source image 'Sea urchin' and crafted adversarial examples for different query budgets by using CGBA for non-targeted attacks. While the adversarial samples crafted against ResNet50 are misclassified as 'Rock-crab', the obtained perturbations against VGG16 and ViT are misclassified as 'Lionfish'. Moreover, the second row of each of the sub-figures depicts amplified (10 times) perturbations for different query budgets. Likewise, Figure~\ref{fig:visul_adv_tar} depicts the adversarial examples and corresponding perturbations of 'Lionfish", crafted by CGBA-H, that are misclassified as 'Sea urchin' by different classifiers. From Figures~\ref{fig:visul_adv_non_tar} and \ref{fig:visul_adv_tar}, we can visualize how the perturbations diminish with the increase of queries. Furthermore, Figures~\ref{fig:non_tar_pert_diff} and \ref{fig:tar_pert_diff} show the difference of the obtained amplified perturbations between different classifiers. From these figures, it is observed that the crafted perturbations vary with classifiers. Because of this variation in crafted perturbations from one classifier to another, the obtained adversarial image in one classifier is not directly transferable to another in case of a decision-based attack.

\end{document}